\RequirePackage{fix-cm}
\documentclass[twocolumn]{svjour3}          % twocolumn
\smartqed  % flush right qed marks, e.g. at end of proof
\usepackage{local}
\usepackage{graphicx}
\usepackage{times}
\usepackage{amsmath,amssymb}
\usepackage{booktabs}
\usepackage{tikz}
\usepackage{cite}
\def\httilde{\hspace{-1.2pt}\mbox{\tt\raisebox{-.9ex}{\large \symbol{126}}}\hspace{-1pt}}

\DeclareMathAlphabet{\mathpzc}{T1}{pzc}{m}{n}
\newlength{\halfwidth}
\newlength{\fullwidth}
\usetikzlibrary{shapes}
\pgfdeclarelayer{background}
\pgfsetlayers{background,main}
\newlength{\tikzimgheight}
\newlength{\tikzimgwidth}

\usepackage{ifthen}
\usetikzlibrary{calc}

\journalname{International Journal of Computer Vision}
\begin{document}

\title{Behavior Discovery and Alignment of Articulated Object Classes from Unstructured Video%\thanks{Grants or other notes
}


\author{Luca Del Pero         \and
        Susanna Ricco \and
        Rahul Sukthankar \and
        Vittorio Ferrari %etc.
}


\institute{Luca Del Pero, \ Vittorio Ferrari \at
              University of Edinburgh, IPAB, School of Informatics \\
              Crichton street 10, Edinburgh EH8 9AB, UK\\
              \email{\{ldelper,vferrari\}@staffmail.ed.ac.uk}
           \and Susanna Ricco, \ Rahul Sukthankar \at
              Google, 1600 Amphitheatre Pkwy \\
              Mountain View, CA 94043, USA \\
           \email{\{ricco,sukthankar\}@google.com}
}

\date{Received: 24 September 2015 / Accepted: 19 July 2016}

\maketitle

\begin{abstract}
We propose an automatic system for organizing the content of a collection 
of unstructured videos of an articulated object class (\eg tiger, horse).
By exploiting the recurring motion patterns of the class across videos,
our system: 1) identifies its characteristic behaviors; and 2)
recovers pixel-to-pixel alignments across different instances.
Our system can be useful for organizing video
collections for indexing and retrieval. Moreover,
it can be a platform for learning the appearance or
behaviors of object classes from Internet video. Traditional
supervised techniques cannot exploit this wealth of data
directly, as they require a large amount of time-consuming manual annotations.

%Internet videos provide a wealth of data that could be
%used to learn the appearance or expected behaviors of many
%object classes. However, most supervised methods cannot exploit this data
%directly, as they require a large amount of time-consuming manual annotations.
%As a step towards solving this problem, 

The behavior discovery stage generates temporal video intervals, 
each automatically trimmed to one instance of the discovered
behavior, clustered by type. It relies on our novel
motion representation for articulated motion based on the
displacement of ordered pairs of trajectories (PoTs).
The alignment stage aligns hundreds of instances of the class to a great
accuracy despite considerable appearance variations (\eg an adult tiger and a cub).
It uses a flexible Thin Plate Spline deformation model that can vary through time.
We carefully evaluate each step of our system on a new, fully annotated dataset.
On behavior discovery, we outperform the state-of-the-art Improved DTF  
descriptor. 
On spatial alignment, we outperform the popular SIFT Flow algorithm.

\end{abstract}

\section{Introduction}
\label{sec:intro}

\begin{figure*}[t]
\begin{center}
\includegraphics[scale =0.37]{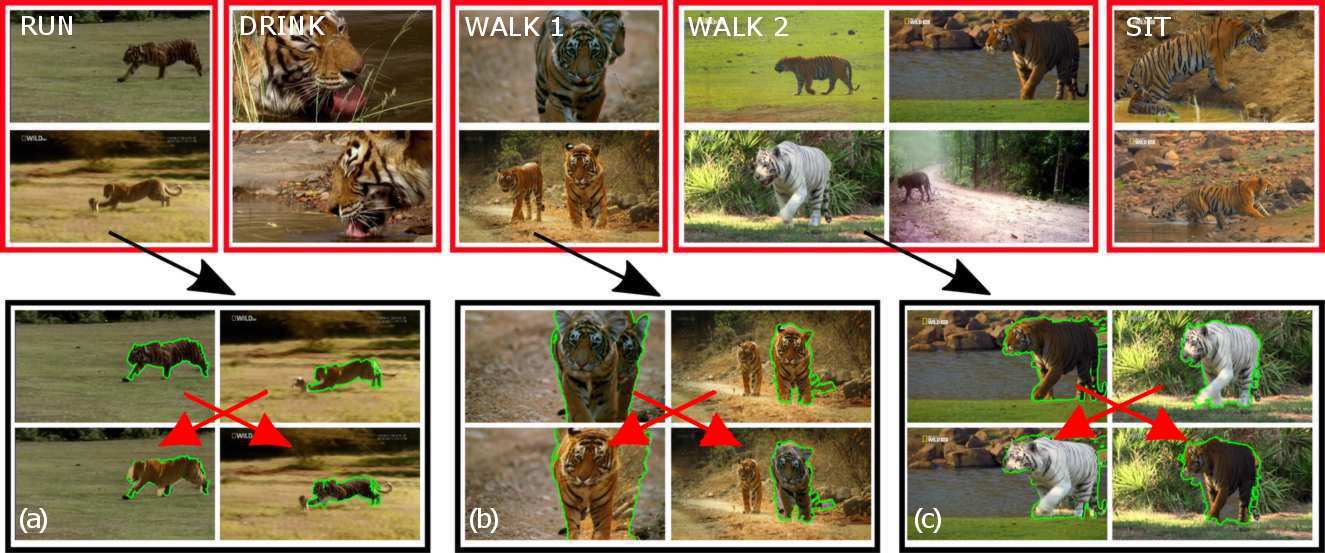}
\end{center}
 \caption{\small{
Output of our system. 
\emph{Top:} examples of the behaviors discovered automatically
from a collection of unstructured videos of an object class (tiger). 
From left to right: running, drinking, two different types 
of walking, and sitting down. Our system uses a new descriptor for articulated motion that analyzes the
displacement of pairs of trajectories. It is fully automatic: the class label is the only
supervision we require. Videos with more behaviors are available at~\cite{delpero15cvpr-potswebpage}.
\emph{Bottom:} within each type of behavior, we find pairs of short sequences 
where the foreground moves in a consistent manner (third row), and align them to a great accuracy 
(fourth row). Here we show an alignment example from the running behavior (a), and one 
each from the two different types of walking (b and c).
The use of motion enables aligning
instances despite large variations in appearance (e.g., white and orange tigers,
adults and cubs). This step is also fully automatic.
}}
\label{fig:teaser}
\end{figure*}

Our goal is to automatically organize the content of a collection of unstructured 
videos of an articulated object class under weak supervision, i.e. only knowing that the object class actually appears in each video (\eg tiger, fig.~\ref{fig:teaser}).
The main contribution of this paper is a fully automatic system that inputs videos of an articulated object
class and discovers its characteristic behaviors (\eg running, walking, sitting down, fig.~\ref{fig:teaser} top), and also recovers the spatial alignment across different instances of the same behavior (fig.~\ref{fig:teaser}, bottom).

%Most computer vision systems cannot directly take advantage of the abundance of Internet video content as training data.
%Traditional supervised methods for learning models of object classes from still 
%images~\cite{Cootes1998ECCV,Felzenszwalb03pictorialstructures,bourdev09iccv,felzenszwalb10pami,wang13iccv,girshick14cvpr} 
%do not easily transfer due to the prohibitive cost of generating ground-truth location annotations in videos. This is especially true for articulated object classes, for which detailed models typically require manual annotation of each part position~\cite{Felzenszwalb03pictorialstructures,bourdev09iccv,yang13pami}.
%
%Analogously, most supervised methods for action recognition 
%cannot exploit Internet videos directly, as they are trained on clips of human
%actors performing scripted actions~\cite{MSRActions,UTInteraction,KTH,WeizmannActions}, 
%usually trimmed to contain a single action~\cite{HMDB,UCF101}.
%Instead, Internet video is typically unstructured and unscripted (\eg clips of animals in the wild). 
%In order to realize the full potential of this vast resource, we must instead rely on methods 
%that require as little manual supervision as possible.

%In general, most computer vision systems 
%cannot directly take advantage of the abundance of Internet video content as training data.

Organizing unstructured video
is important for a wide variety of applications, such as video indexing and retrieval (\eg the TRECVid conference series~\cite{smeaton06mir}), video database summarization (\eg ~\cite{tompkin12siggraph,wang09icme}), and computer graphics applications (\eg replacing an instance in a video with one from a different video, fig.~\ref{fig:teaser}). Moreover, it can help generate training data for supervised systems
for action recognition (\eg ~\cite{wang_ICCV_2013,MSRActions,KTH,WeizmannActions}) and object class detection
(\eg ~\cite{Felzenszwalb03pictorialstructures,bourdev09iccv,felzenszwalb10pami,wang13iccv,girshick14cvpr}). 
These methods cannot fully exploit the abundance of unstructured Internet videos due to the prohibitive cost of generating ground-truth annotations, which explains the recent interest in
learning from video under weak supervision~\cite{Leistner11,prest12cvpr,Tang2013}.

Our method requires very little supervision (one class label per video), and could
potentially replace the tedious and time-consuming manual annotations needed by supervised recognition systems.
For example, action recognition systems are typically trained on clips of human actors performing scripted actions~\cite{MSRActions,UTInteraction,KTH,WeizmannActions}, 
usually trimmed to contain a single action~\cite{HMDB,UCF101}. 
Discovering the behaviors of a class from unstructured video could replace the process 
of searching for examples of each behavior, as well as temporally segmenting them out of the videos by hand.
Analogously, traditional supervised methods for learning models of object classes from still images~\cite{Cootes1998ECCV,Felzenszwalb03pictorialstructures,bourdev09iccv,felzenszwalb10pami,wang13iccv,girshick14cvpr} 
do not easily transfer to videos as they require expensive location annotations.
The alignments recovered by our method could potentially
replace the manual correspondences needed by most popular methods for learning object classes~\cite{dalal05cvpr,felzenszwalb10pami,viola:nips05,cinbis13iccv,wang13iccv,girshick14cvpr},
including those requiring part-level annotations~\cite{Felzenszwalb03pictorialstructures,bourdev09iccv,azizpour12eccv}.
They can also enable annotating large collections with little manual
effort via knowledge transfer~\cite{vezhnevets14cvpr,kuettel12eccv,lampert:cvpr09,fei2007CVIU,MalisiewiczICCV11}.
One could provide manual annotations only for a few instances (\eg segmentation masks~\cite{MalisiewiczICCV11,vezhnevets14cvpr,kuettel12eccv} or 3D models~\cite{MalisiewiczICCV11,tighe13cvpr}), and then propagate
them automatically to many more instances via the recovered alignments.

Our focus is on highly articulated, deformable objects like animals. 
Such classes are typically challenging, 
as they exhibit a much wider variety of interesting behaviors compared to more rigid objects (\eg a train).
Moreover, aligning such objects is challenging due to their deformable nature.
These are also the reasons why articulated classes typically require a greater annotation effort than rigid ones~\cite{Felzenszwalb03pictorialstructures,bourdev09iccv,yang13pami}.

A preliminary version of this work appeared at CVPR 2015~\cite{delpero15cvpr}
covering the behavior discovery stage. In this journal paper, we introduce the
spatial alignment stage and present a more extensive experimental evaluation.

\section{Overview of our approach}
\label{sec:overview}

Given unstructured videos of an articulated object class, we discover the class behaviors
and recover spatial alignments across different class instances (fig.~\ref{fig:teaser}).
We exploit the nature of video and recent advances in motion 
analysis~\cite{papazoglou13iccv,Wang_2011_CVPR,wang_ICCV_2013} to
make our system fully automatic, for example we use motion segmentation~\cite{papazoglou13iccv} 
to estimate the object's 2-D location.

To model the motion of an articulated object, we introduce
a new descriptor that captures the relative motion of its parts,
for example the knee and the ankle of an animal walking.
We do this by analyzing the relative displacement of Pairs
of point Trajectories (PoTs). PoTs are a key component of this work,
which we discuss and evaluate in detail.

Our system consists of two main stages: behavior discovery and spatial alignment.
The behavior discovery builds on the PoTs descriptor.
For the spatial alignment, we introduce a technique for aligning short video sequences of the same
object class based on Thin Plate Splines (TPS).

\paragraph{Pairs of Trajectories (sec.~\ref{sec:pots}).}
We model articulated motion by analyzing the relative
displacement of large numbers of ordered trajectory pairs (PoTs).
The first trajectory in the pair defines a reference frame in which the motion of the second
trajectory is measured.  We preferentially sample pairs across
joints, resulting in features particularly well-suited to representing
the behavior of articulated objects. This has greater
discriminative power than state-of-the-art features defined using single
trajectories in isolation~\cite{Wang_2011_CVPR,wang_ICCV_2013}.

In contrast to other popular descriptors~\cite{Jain2013,Wang_2011_CVPR,wang_ICCV_2013}, 
PoTs are defined solely by motion and so are robust to appearance variations within the object class. In cases
where appearance proves beneficial for discriminating between behaviors of
interest, it is easy to combine PoTs with standard appearance features.

\begin{figure*}[t]
\begin{center}
\includegraphics[scale =0.39]{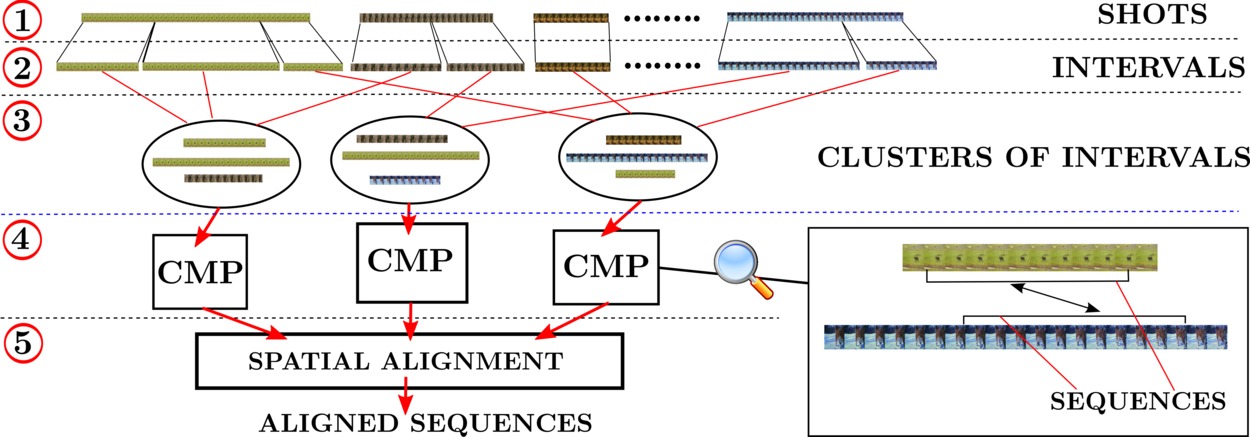}
\end{center}
 \caption{\small{
System architecture and terminology (sec.~\ref{sec:architecture}). 
The input is a collection of shots showing the
same class (top), which can be of any length. 
We first extract foreground masks~\cite{papazoglou13iccv}.
We then extract PoTs descriptors from each shot (step~1, sec.~\ref{sec:pots}).
Each shot is then partitioned into shorter temporal
intervals that are each likely to contain a single behavior (step~2, sec.~\ref{sec:timepartitioning}), 
which we cluster using PoTs (step~3, sec.~\ref{sec:timeclustering}).
For each two intervals in the same cluster, 
we extract pairs of short sequences showing consistent foreground motion (CMPs), which become
candidates for spatial alignment (step~4, sec.~\ref{sec:cmpcandidates}). 
Last, we align the two sequences of each CMP (step~5, sec.~\ref{sec:homography} and \ref{sec:tps}).
}}
\label{fig:architecture}
\end{figure*}

\paragraph{Behavior discovery (sec.~\ref{sec:discovery}).}
Our method does not require knowledge of the number or
types of behaviors, nor that instances of different behaviors be temporally
segmented within a video.
Instead, we leverage that behaviors exhibit consistency
across videos, resulting in characteristic motion patterns.
Our method identifies motion patterns that recur across several videos:
it temporally segments them out of the input videos, and clusters them
by type. For this, we use PoTs as motion representation, which allow us to distinguish
between fine-grained behaviors, such as walking and running.
Note that our \emph{unsupervised} discovery is very different from simply classifying fixed temporal 
chunks of video into behaviors (\eg, action recognition in UCF-101~\cite{UCF101}), 
which requires \emph{supervision} 
(\eg training data for each behavior), and does not need to
address the temporal segmentation.

\paragraph{Spatial alignment (sec.~\ref{sec:alignment}).}
Consider the problem of aligning any two instances of a tiger.
This is challenging due to differences in viewpoint (\eg frontal
and side), pose (jumping and sitting down), and appearance (cub and adult).
The behavior discovery stage simplifies the problem by forming clusters
of videos exhibiting a consistent set of poses (\eg walking, jumping). 
However, aligning two individual frames
with traditional techniques for aligning still images~\cite{barnes10eccv,liu08eccv,Hartley00,lowe04ijcv} 
typically fails even in this scenario, due to the significant appearance variations across instances and pose variations within the same behavior (\eg different phases of walking).
Instead, we align two short temporal sequences where the objects exhibit consistent motion (we identify these sequences automatically within the behavior clusters).

We exploit the consistency in object motion to establish reliable point correspondences between the
sequences, and combine this with edge features to align them with great accuracy (fig.~\ref{fig:teaser}).
We model the transformation between the two sequences using a
series of Thin Plate Splines (TPS)~\cite{wahba90tps}.
TPS are an expressive non-rigid mapping that can accommodate for the deformations of complex articulated objects. 
TPS have been used before mostly for registration~\cite{Chui03} and shape matching~\cite{ferrari10ijcv} in still images.
We extend these ideas to video by fitting TPS that vary smoothly in time.

\subsection{System architecture}
\label{sec:architecture}
We provide here a high-level description of the architecture of our system
(fig.~\ref{fig:architecture}).

\paragraph{Input video shots.}
The input is a collection of \emph{video shots} of the same object class.
By shot we mean a sequence of frames without scene transitions~\cite{kim09isce}.
We work with Internet videos automatically partitioned into shots by thresholding histogram differences across consecutive frames~\cite{prest12cvpr,kim09isce}.
The only supervision given is the knowledge that each shot contains the object class.

\paragraph{Foreground masks.}
We use the fast video segmentation technique~\cite{papazoglou13iccv} on each input shot, 
to automatically segment the foreground object from the background. 
These foreground masks remove features on the background and facilitate
the entire process.
To handle shots containing multiple moving objects,
we only keep the largest connected component in the foreground mask.
This typically corresponds to the largest object in the shot (a similar
strategy is used in~\cite{papazoglou13iccv} for evaluation).

\paragraph{Step 1: PoT extraction (sec.~\ref{sec:pots}).}
We extract PoTs from each input shot, which we use as features in the following
stjpg.

\paragraph{Step 2: Partitioning into temporal intervals (sec.~\ref{sec:timepartitioning}).}
Clustering the input shots directly would fail to discover behaviors,
since each shot typically contains several different behaviors.
For example, a tiger may walk for a while, then sit down and finally stretch.
We use motion cues to partition shots into \emph{single-behavior intervals}, 
\eg, a ``walking'', a ``sitting down'' and a ``stretching'' interval.

\paragraph{Step 3: Behavior discovery by clustering (sec.~\ref{sec:timeclustering}).}
We use the extracted PoTs to build a descriptor for each
interval from step 2, and cluster them.
At this stage of the pipeline, each cluster contains several intervals of
the same behavior, each temporally trimmed to its duration.

\paragraph{Step 4: Candidates for Spatial Alignment (sec.~\ref{sec:cmpcandidates}).}
We exploit the consistent motion of two intervals in the same
behavior cluster to drive their alignment.
However, we cannot expect the motion to be consistent for their entire duration:
this would require that the object performs exactly the same movements in the same order in both intervals.
Hence, we identify a few shorter sequences of fixed length
between the two intervals
that exhibit consistent foreground motion. We term these \emph{Consistent Motion Pairs} (CMPs),
and use them as candidates for spatial alignment.

\paragraph{Step 5: Spatial alignment (sec.~\ref{sec:homography},~\ref{sec:tps}).}
For each CMP, we attempt to align its two sequences. 
If the algorithm succeeds, we output the aligned CMP.
We consider two different spatial alignment models:
homographies and TPS.

\subsection{Experiments overview}
\label{sec:overviewexperiments}
We present extensive quantitative evaluation on a new dataset 
containing several hundreds videos of three articulated object classes 
(dogs, horses and tigers, sec.~\ref{sec:dataset}). 
We produced the annotations necessary to evaluate the two outputs 
of our method:
1) per-frame behavior labels in over 110,000 frames to evaluate behavior discovery; and
2) 2-D positions of $19$ landmarks (e.g. left eye, front right ankle) in over
35,000 frames to evaluate spatial alignment.

The results demonstrate that our method can discover behaviors from a collection
of unconstrained video, while also segmenting out behavior
instances from the input videos (sec.~\ref{sec:potsevaluation} and \ref{sec:discoveryevaluation}). 
On these tasks, PoTs perform significantly better than
existing appearance- and trajectory-based descriptors (e.g., HOG and DTFs~\cite{wang_ICCV_2013}). 

Our TPS based alignment outperforms existing alternatives
that are either unsuitable for articulated objects (\eg homographies~\cite{caspi06ijcv,Hartley00,lowe04ijcv}), 
or designed to align still images (\eg the popular SIFT Flow algorithm~\cite{liu08eccv}). 
Our system recovers approximately 1000 pairs of correctly aligned sequences
from 100 real-world video shots of tigers, and $800$ aligned sequences from $100$ shots of horses.
As the recovered alignment is between {\em sequences}, this entails correspondences 
between several thousand pairs of frames (sec.~\ref{sec:resultsalignment}).

\begin{figure*}[t]
\begin{center}
\includegraphics[scale =0.48]{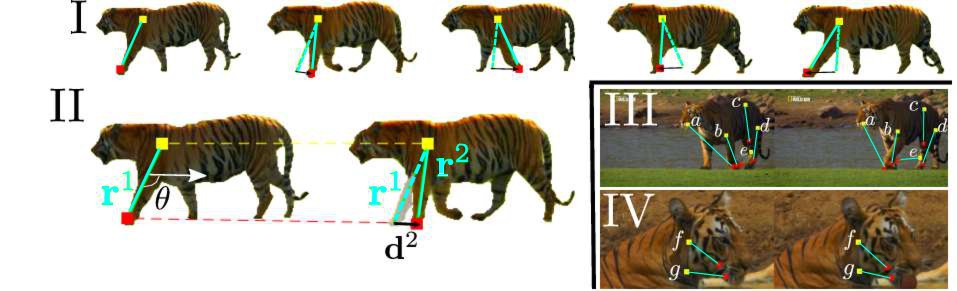}
\end{center}
 \caption{\small{
Modeling articulated motion with PoTs.
Two trajectories in a PoT are ordered  based on their deviation from the median
velocity of the object: the anchor (yellow) deviates less than the swing (red).
In I, the displacement of the swing relative to the anchor follows the swinging
motion of the paw with respect to the shoulder. While both move forward as the
tiger walks, the paw is actually moving backwards in a coordinate system
centered at the shoulder. This back-and-forth motion is captured by the
relative displacement vectors of the pair (in black) but missed when individual
trajectories are used alone. The PoT descriptor is constructed from the angle
$\theta$ and the black vectors $\mathbf{d^k}$, shown in II (sec.~\ref{sec:potdefinition}).  
The two trajectories in a PoT are selected such that they track object parts that 
move differently. A few selected PoTs are shown in III and IV. Paws move
differently than the head (a), hip (c), knees (b,d), or other paws (e). In IV,
the head rotates relative to the neck, resulting in different PoTs (f,g).  Our
method selects these PoTs without requiring prior knowledge of the object
topology (sec.~\ref{sec:potselection}).
 } } \label{fig:pots} \end{figure*}

\section{Pair of trajectories (PoTs)}
\label{sec:pots}

\begin{figure*}[t]
\begin{center}
\setlength{\tabcolsep}{0.8pt}
\begin{tabular}{c c c c c c}
\footnotesize{\textbf{input foreground}} & \footnotesize{\textbf{deviation from}} & \footnotesize{\textbf{extracted PoTs}} & \footnotesize{\textbf{anchors and swings}}  & \footnotesize{\textbf{extracted PoTs}} & \footnotesize{\textbf{anchors and swings}} \\[-3pt]
\footnotesize{\textbf{trajectories}} & \footnotesize{\textbf{median velocity}} & \scriptsize{$\mathbf{(\theta_{P}=0.01)}$} & \scriptsize{$\mathbf{(\theta_{P}=0.01)}$}  & \scriptsize{$\mathbf{(\theta_{P}=0.15)}$} & \scriptsize{$\mathbf{(\theta_{P}=0.15)}$} \\
\includegraphics[scale = 0.2]{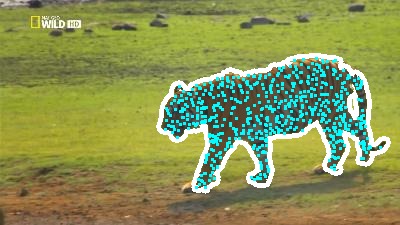}  &
\includegraphics[scale = 0.2]{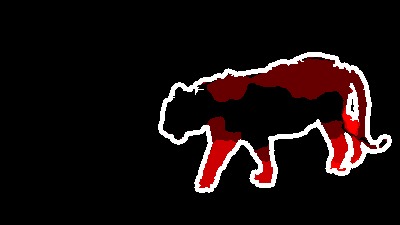}  &
\includegraphics[scale = 0.2]{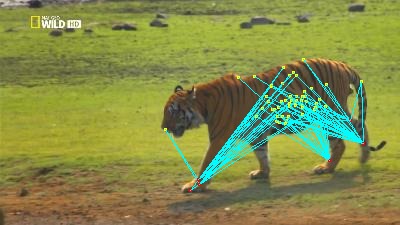}  &
\includegraphics[scale = 0.2]{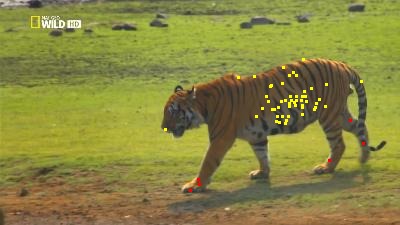}  &
\includegraphics[scale = 0.2]{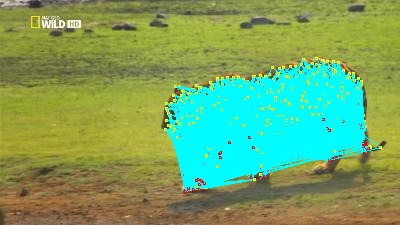}  &
\includegraphics[scale = 0.2]{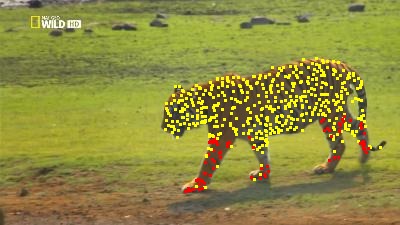} \\
\includegraphics[scale = 0.2]{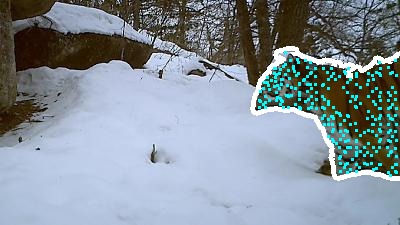}  &
\includegraphics[scale = 0.2]{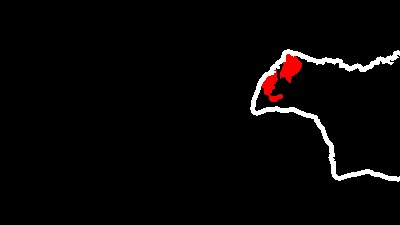}  &
\includegraphics[scale = 0.2]{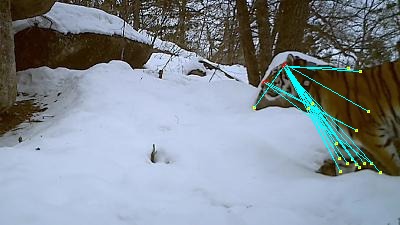}  &
\includegraphics[scale = 0.2]{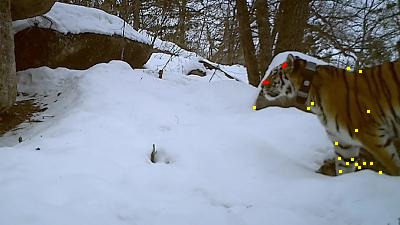}  &
\includegraphics[scale = 0.2]{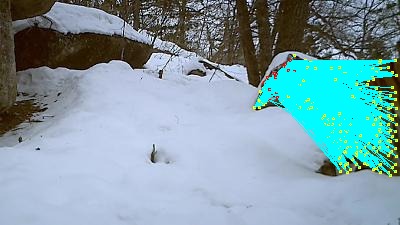}  &
\includegraphics[scale = 0.2]{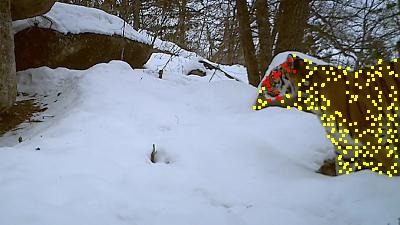} \\
\small{(a)} & \small{(b)} & \small{(c)} & \small{(d)} & \small{(e)} & \small{(f)} \\
\end{tabular}
\end{center}
 \caption{\small{
PoT selection on two different examples: a tiger walking (top) and one turning
its head (bottom). We construct PoT candidates from the trajectories on the
foreground mask (a), using all possible pairs (sec.~\ref{sec:potselection}). 
We prefer candidates where the
anchor is closer to the median foreground velocity, denoted by dark areas in (b), while the
swing follows a different motion (bright areas). We keep the highest
$\theta_{P} \%$ ranking candidates according to this criterion. We show the
selected PoTs for two different values of $\theta_{P}$ (c,e).
Too strict a $\theta_{P}$ ignores many interesting PoTs (c), like those involving trajectories
on the head in the top row.
We also show the trajectories used as anchors (yellow)  and swings (red) without the lines connecting them 
(d,f). Imagine connecting any anchor with any
swing: in most cases, the two follow different, independently moving parts of
the object, which is the key requirement of a PoT. 
We use $\theta_{P} = 0.15$ in our experiments (e,f).
}
}
\label{fig:potselection} \end{figure*}

We represent articulated object motion using a collection of
ordered pairs of trajectories (PoTs), tracked over $n$ frames.
We compose PoTs from the trajectories extracted with 
a dense point tracker (\eg ~\cite{wang_ICCV_2013}):
only two trajectories following parts of the object moving relatively to each other are selected 
as a PoT, as these are the pairs that move in a consistent and distinctive manner across 
different instances of a specific behavior.
For example, the motion of a pair connecting a tiger's knee to
its paw consistently recurs across videos of walking tigers (figs.~\ref{fig:pots} and~\ref{fig:potselection}).
By contrast, a pair connecting two points on the chest (a
rather rigid structure) may be insufficiently distinctive,
while one connecting the tip of the tail to the nose may lack consistency.
Note also that a trajectory may simultaneously contribute to multiple
PoTs (\eg, a trajectory on the front paw may form pairs with trajectories from
the shoulder, hip, and nose).

Although we often refer to PoTs using semantic labels for the location of their
component trajectories (eye, shoulder, hip, etc.), these are used only for
convenience.  PoTs do not require semantic understanding or any part-based or
skeletal model of the object, nor are they specific to a certain object class.
Furthermore, the collection of PoTs is more expressive than a simple star-like
model in which the motion of point trajectories are measured relative to the
center of mass of the object (\ie, normalizing by the dominant object motion). 
For example, we find the ``walking" cluster (fig.~\ref{fig:teaser}) 
based on PoTs formed by various combinations
of  head-paw (fig.~\ref{fig:pots} III, a), hip-knee (c), knee-paw (b,d),
or even paw-paw trajectories (e). 

Fig.~\ref{fig:pots} (III-IV) shows a few examples of PoTs selected from two tiger videos. We define PoTs 
in sec.~\ref{sec:potdefinition}, while we explain how to select PoTs from real videos in sec.~\ref{sec:potselection}.

\subsection{PoT definition}
\label{sec:potdefinition}
\paragraph{PoT ordering: anchors and swings.} 
The two trajectories in a PoT are \emph{ordered}, \ie
we always measure the displacement of the second trajectory (\emph{swing}) in
the local coordinate frame defined by the first (\emph{anchor}).
%the first trajectory in a PoT (\emph{anchor})
%defines a local coordinate frame in which the motion of the second
%(\emph{swing}) is measured.
We select as anchor the trajectory whose velocity is closer to the median velocity of 
pixels on the foreground mask, aggregated over the length of the PoT (sec.~\ref{sec:potselection}). This approximates the median velocity of the whole object.
This criterion generates a stable ordering, repeatable across the broad range of videos we examine.
For example, the trajectories on the legs in fig.~\ref{fig:pots}
are consistently chosen as swings while those on the torso as anchors.

\paragraph{Displacement vectors.}
In each frame $f_k$, we compute the vector $\mathbf{r^{k}}$
from anchor to swing (cyan lines in fig.~\ref{fig:pots}).
Starting from the second frame, a displacement vector $\mathbf{d^{k}}$ 
is computed by subtracting the vector $\mathbf{r^{k-1}}$
of the previous frame (dashed cyan) from the current $\mathbf{r^{k}}$ (solid cyan).
$\mathbf{d^k}$ captures the motion of
the swing relative to the anchor by canceling out the motion of the latter. 
Naively employing the cyan vectors $\mathbf{r^k}$ as raw features
does not capture relative motion as effectively, because
the variation in $\mathbf{r^k}$ through time is dominated by the
spatial arrangement of anchor and swing rather than by the change in relative
position between frames (compare the magnitudes of the cyan
and black vectors in fig.~\ref{fig:pots}).
Note this way of computing the displacement vectors is invariant to camera panning, since the relative motion of the trajectories does not change whether the camera is static or panning.

\paragraph{PoT descriptor.}
The descriptor $P$ has two parts:
1) the initial position of the swing relative to the anchor, and
2) the sequence of normalized displacement vectors over time:
\begin{equation}
P=\left(
	\theta,
	\frac{\mathbf{d^2}}{D}, \ldots, \frac{\mathbf{d^n}}{D}
  \right),
\label{eq:potdescriptor}
\end{equation} 
where $\theta$ is the angle from anchor to swing in the first frame (in radians)
and the normalization factor is the total displacement $D=\sum_{k=2}^{n}||\mathbf{d^{k}}||$.
The dense trajectory features (DTF) descriptor~\cite{Wang_2011_CVPR} employs a similar normalization.
Note also that the first term in $P$ records only the angle (and not
the magnitude) between anchor and swing; this retains scale invariance and
enables matching PoTs between objects of different size.
The dimensionality of $P$ is $2\cdot (n-1)+1$; in all of our experiments $n = 10$.

\subsection{PoT selection}
\label{sec:potselection}

We explain here how to automatically form PoTs out of a set of trajectories extracted with a dense point tracker~\cite{wang_ICCV_2013}. 
We start with a summary of the process and give more details later.
First, we remove trajectories on the background using the foreground masks.
Then, for each frame $f$ we build the set $\mathcal{P}_{f}$ of PoTs starting at that frame.
For computational efficiency, we directly set $\mathcal{P}_{f}=\emptyset$ for any frame unlikely to contain articulated motion. 
Otherwise, we form candidate PoTs from all pairs of 
foreground trajectories $\{t_i,t_j\}$ extending for at least $n$ frames after $f$.
Finally, we retain in $\mathcal{P}_{f}$ the candidates most likely to be on object parts moving relative to each other.

\paragraph{Removing background trajectories.}
State-of-the-art point trajectories already attempt to limit trajectories to
foreground objects~\cite{wang_ICCV_2013}, but often fail on the wide range of videos we use.
The video segmentation technique we use~\cite{papazoglou13iccv} handles unconstrained video, and reliably detects articulated objects even under significant motion and against cluttered backgrounds.
Hence, we remove point trajectories that fall outside the foreground mask produced by~\cite{papazoglou13iccv}.
Results show that
our overall method is robust to inaccurate foreground masks because they only affect a fraction of the PoT collection (sec.~\ref{sec:potsevaluation}).

We also use the masks to estimate the median velocity of the object, computed as the median optical flow displacement over all pixels in the mask.

\paragraph{Pruning frames without articulated motion.}
A frame is unlikely to contain articulated motion (hence PoTs) if the optical
flow displacement of foreground pixels is uniform.
This happens when the entire scene is static,
or the object moves with respect to the camera but the motion is not articulated. 
We define
$s(f) = \frac{1}{n}\sum_{i=f}^{f+n-1}\sigma_{i}$,
where $\sigma_{i}$ is the standard deviation in the optical flow displacement over
the foreground pixels at frame $i$ normalized by the mean, and $n$ the length of the PoT.
We set $\mathcal{P}_{f}=\emptyset$ for all frames where
$s(f) < \theta_{F}$, thereby pruning frames unlikely to contain any PoT.
We choose $\theta$ on $16$ cat videos in which we manually labeled frames without articulated motion.
We set $\theta_{F}=0.1$, which yields precision $0.95$ and recall $0.75$ (very similar performance
is achieved for $0.05 \leq \theta_{F} \leq 0.2$).

\paragraph{PoT candidates and selection.}
The candidate PoTs for a
frame $f$ are all ordered pairs of trajectories $\{t_i,t_j\}$ 
that start in $f$ and exist in the following $n-1$ frames (fig.~\ref{fig:potselection}a).
We score a candidate pair $\{t_i,t_j\}$ using
\begin{equation}
  \mathrm{S}(\{t_i=a, t_j=s\}) = \sum_{k=f}^{f+n-1}||v_{s}^{k} - v_{m}^{k}|| - ||v_{a}^{k} - v_{m}
^{k}||~~,
\label{eq:potscore}
\end{equation}
where $v_{m}^{k}$ is the median velocity at frame $k$, and
$v_{s}^{k},v_{a}^{k}$ are the velocities of 
the swing and anchor.
The first term favors pairs where the swing velocity deviates a lot from the median, while the second term favors pairs where the anchor velocity is close to the median.
As seen in fig.~\ref{fig:potselection}, this generates a stable PoT ordering, for example the swings fall
on the legs as the tiger walks (top), or on the turning head (bottom).
We rank all candidates using (\ref{eq:potscore})
and retain the top $\theta_{P} \%$ candidates as PoTs $\mathcal{P}_{f}$ for this frame (fig.~\ref{fig:potselection}c-f).
In all experiments we use $\theta_{P} = 0.15$.
Since we score all possible pairs with (\ref{eq:potscore}),
a particular trajectory can serve as anchor in one pair
and as swing in a different pair, depending
on the velocity of the other trajectory in the pair.
%This is much more robust and flexible
%than partitioning trajectories into anchors and swings and then 
%only considering  pairs across the two sets.}

\section{Behavior discovery}
\label{sec:discovery}

The behavior discovery stage inputs a set of shots $\mathcal{S}$ of the same class (fig.~\ref{fig:architecture}, top) and outputs clusters of temporal intervals, $\mathcal{C}=(c_{1},...,c_{k})$ corresponding to behaviors 
(step 3 in fig.~\ref{fig:architecture}). 
For the ``tiger'' class, we would like a cluster with tigers
walking, one with tigers turning their head, and so on.
We first temporally partition shots into single behavior intervals (sec.~\ref{sec:timepartitioning}).
Then we cluster these intervals to discover recurring behaviors (sec.~\ref{sec:timeclustering}).

\subsection{Temporal partitioning}
\label{sec:timepartitioning}
An input shot typically contains several instances of 
different behaviors each.
It would be easier to cluster intervals corresponding to just one instance of a behavior, and ideally covering its whole duration.
Here we partition each shot into such single behavior intervals.
Boundaries between such intervals cannot be detected using simple color histogram differences (unlike shot boundaries~\cite{kim09isce}).
Further, naively partitioning into fixed-length intervals invariably ends up either over- or under-partitioning.
Instead, we use an adaptive strategy based on two different motion cues: pauses and periodicity.

\paragraph{Partitioning on pauses.}
The object often stays still for a brief moment between two different behaviors.
We detect such pauses as sequences of three or more frames without articulated
object motion (sec.~\ref{sec:potselection}).

\paragraph{Partitioning based on periodicity.}
As some sequences lack pauses between different, but related behaviors (\eg, from walking to running), we also partition based on periodic motion.
For this we use time-frequency analysis, as
periodic motion patterns like walking, running, or licking
typically generate peaks in the frequency domain (examples
available on our website~\cite{delpero15cvpr-potswebpage}). 

We model an interval as a time sequence 
$s(t)=b_{f^{t}}^{P}$, where $b_{f^{t}}^{P}$ is
a bag-of-words (BoW) of PoTs at frame $f^{t}$. 
We convert $s(t)$ to $V$ one-dimensional sequences
and sum the Fast Fourier Transform (FFT) of the individual sequences in the frequency domain
($V$ is the codebook size).
If the height of the highest peak is $\geq \theta_{H}$, 
we consider the interval as periodic.
We normalize the total energy to make sure it integrates to $1$.
Using the sum of the FFTs makes the approach more
robust, since peaks arise only if several codewords recur with the
same frequency. 

Naively doing time-frequency analysis on an entire interval
typically fails because it might contain both periodic
and non-periodic motion (\eg, a tiger walks for a while
and then sits down). Hence, we consider all possible sub-intervals using
a temporal sliding window and label the one with the highest 
peak as periodic, provided its height $\geq \theta_{H}$. The remaining segments are reprocessed to extract motion patterns with different periods (\eg, walking versus running) until no significant peaks remain. 
For robustness, we only consider sub-intervals
where the period is at least five frames
and the frequency at least three
(\ie, the period repeats at least three times).
We empirically set $\theta_{H} = 0.1$, which
produces very few false positives.

\vspace{-8pt}
\subsection{Clustering intervals}
\label{sec:timeclustering}
\vspace{-6pt}

We use $k$-means to form a codebook from one million PoT descriptors randomly sampled from all intervals,
using Euclidean distance\footnote{Since the PoT descriptor is heterogenous (sec.~\ref{sec:potdefinition}), 
we ran preliminary experiments on held-out data to weigh the relative importance of $\theta$ and the displacement
vectors (analogously to the way we set the other parameters, sec.~\ref{sec:potsprotocol}). We found the optimal
weight is 1.}.
We run $k$-means eight times and choose the clustering
with lowest energy to reduce the effects of random 
initialization~\cite{wang_ICCV_2013}.
We then represent an interval as a BoW histogram of the PoTs it contains (L1-normalized).

We cluster the intervals using hierarchical clustering with complete-linkage~\cite{johnson67psychometrica}.
We found this to perform better than other clustering methods (\eg, single-linkage, $k$-means)
for both PoTs and the Improved DTFs (IDTFs)~\cite{wang_ICCV_2013} descriptor,
which we compare against in the experiments (sec.~\ref{sec:potsevaluation}).

\begin{figure}
\begin{center}
\includegraphics[scale = 0.22]{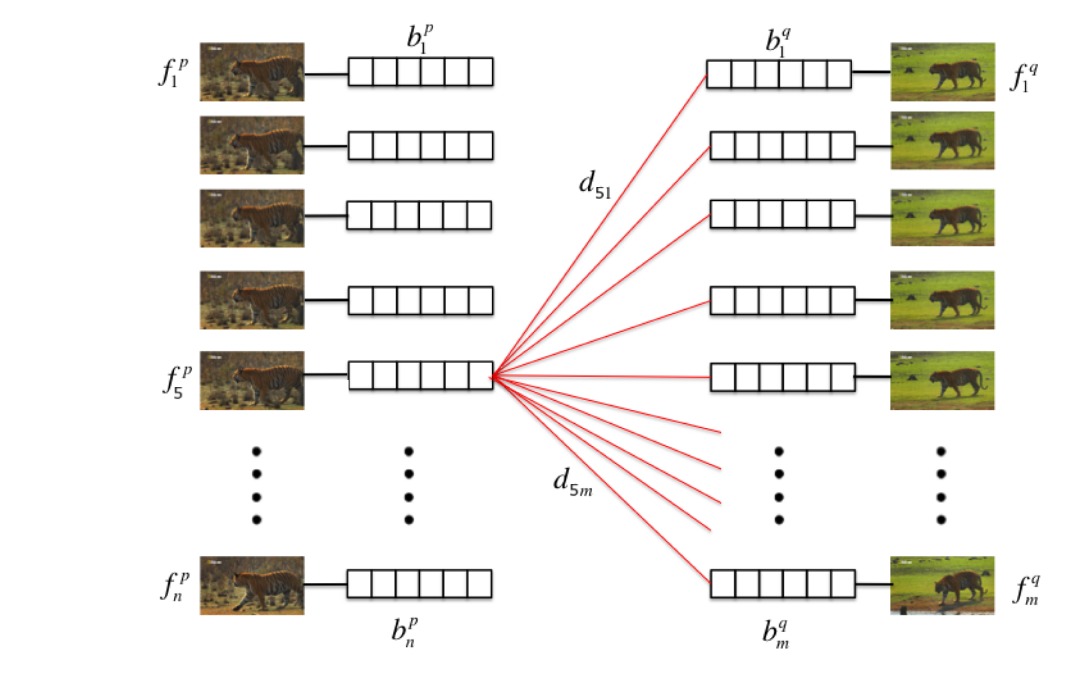}
\end{center}
 \caption{\small{
Extracting CMPs from two intervals.
First, we approximate the pairwise distance between frames
as the histogram distance between their BoWs (which contains all
motion descriptors through the frame, sec.~\ref{sec:cmpcandidates}).
Then we keep as CMPs the top scoring pairs of sequences of length $T$ with respect to
(\ref{eq:cmpcandidates}). For the intervals above, the number of pairs of sequences to score is $(n-T)\cdot (m-T)$.
}}
\label{fig:cmpcandidates}
\end{figure}

\begin{figure*}[t]
\begin{center}
\setlength{\tabcolsep}{0.8pt}
\begin{tabular}{c c c c c c}
\footnotesize{\textbf{foreground masks}} & \footnotesize{\textbf{trajectory matches}} & \footnotesize{\textbf{homography}} & \footnotesize{\textbf{TPS mapping}} &  \footnotesize{\textbf{foreground edge points}} \\
\includegraphics[scale =0.25]{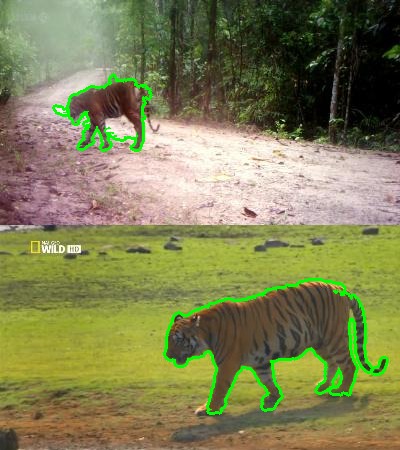}  &
\includegraphics[scale =0.25]{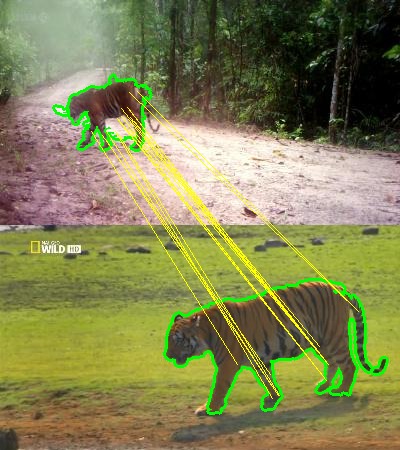}  &
\includegraphics[scale =0.25]{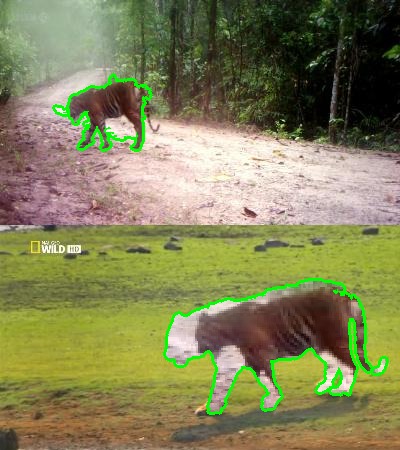}  &
\includegraphics[scale =0.25]{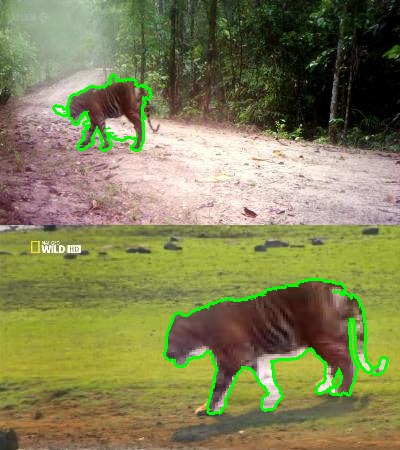}  &
\includegraphics[scale =0.25]{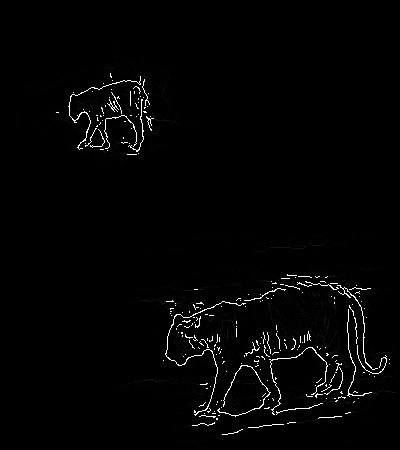}  \\
\small{(a)} & \small{(b)} & \small{(c)} & \small{(d)} & \small{(e)} \\
\end{tabular}
\end{center}
 \caption{\small{
Aligning sequences with similar foreground motion. We first estimate
a foreground mask (green) using motion segmentation (a). We then fit a 
homography to matches between point trajectories (b, sec.~\ref{sec:homography}).
In (c) we project the foreground pixels in the first sequence 
(top) onto the second (bottom) with the recovered homography.
This global, coarse mapping is often not accurate (note the
misaligned legs and head).
We refine it by fitting Thin-Plate Splines (TPS) to
edge points extracted from the foreground (e, sec.~\ref{sec:tps}).
The TPS mapping is non-rigid and provides a more accurate alignment 
for complex articulated objects (d).
}
}
\label{fig:sequencealignment} \end{figure*}

Hierarchical clustering requires computing distances between items. Given BoWs of PoTs $b_{u}$ and $b_{v}$ for intervals $I_{u}$ and $I_{v}$, 
we use
\begin{equation}
\mathrm{d}(I_{u},I_{v})= -\mathrm{exp}\left(~-(1-\mathrm{HI}(b_{u},b_{v})~)~\right) ,
\label{eq:clusterdistance}
\end{equation}
where $\mathrm{HI}$ denotes histogram intersection. We found this
to perform slightly better than the $\chi^2$ distance.
Note that this function can be also used on BoWs of descriptors
other than PoTs.
Additionally, it can be extended to handle different descriptors that use multiple
feature channels, such as IDTFs~\cite{wang_ICCV_2013}.
In this case, the interval representation is a set of BoWs $(b_{u}^{1},...,b_{u}^{C})$, 
one for each of the $C$ channels.
Following~\cite{wang_ICCV_2013},
we combine all channels into a single distance function
\begin{equation}
\mathrm{d}(I_{u},I_{v})= -\mathrm{exp}\left(-\sum_{i=1}^{C} \frac{1-\mathrm{HI}(b_{u}^{i},b_{v}^{i})}{A_{i}}\right) ,
\label{eq:combinedclusteringdistance}
\end{equation}
where $A_{i}$ is the average value of $(1-\mathrm{HI})$ for channel $i$.

\vspace{-8pt}
\section{Sequence alignment}
\label{sec:alignment}

Having clustered intervals by behavior type, we can search for suitable candidates for spatial alignment,
\ie pairs of short sequences with consistent foreground motion (dubbed CMPs, fig.~\ref{fig:architecture} step 4). 
This is discussed in sec.~\ref{sec:cmpcandidates}.

We have explored a variety of approaches for sequence alignment, 
and report on two representative methods here (fig.~\ref{fig:sequencealignment}).
The first is a coarse, global alignment generated by fitting a single homography to foreground trajectory descriptors matched between the two sequences (sec.~\ref{sec:homography}).
The second approach fits a finer, non-rigid TPS mapping to edge points extracted from the foreground regions of each frame.  
We allow TPS to vary smoothly through the sequence (sec.~\ref{sec:tps}).
As we show in our experiments, the TPS prove more suitable for aligning complex articulated objects (sec.~\ref{sec:alignmentevaluation}).

\vspace{-8pt}

\subsection{Extracting CMP candidates}
\label{sec:cmpcandidates}

Given two intervals $p$ and $q$ in the same behavior cluster, 
we extract as CMPs the top 10 ranked pairs of subsequences between them 
according to the following metric (fig.~\ref{fig:cmpcandidates}).
Let $d_{ij}$ be the histogram intersection between BoW descriptors
computed for frame $i$ in $p$ and frame $j$ in $q$. We compute $d_{ij}$ 
just like in (\ref{eq:combinedclusteringdistance}), except that
we aggregate only the descriptors in the specific frame rather than the whole interval.
The similarity between the $T$-frame subsequence of $p$ starting at frame $i$
and the subsequence of $q$ starting at frame $j$ is
\begin{equation}
\tiny{\mathrm{s}\!\left([f_{i}^{p},\ldots,f_{i+T-1}^{p}]  ,  [f_{j}^{q},\ldots,f_{j+T-1}^{q}]\right)=\sum_{t=0}^{T-1}d_{(i+t)(j+t)}}
\label{eq:cmpcandidates}
\end{equation}
This measure preserves the temporal order of the frames, whereas aggregating the BoW
over the whole sequences as in (\ref{eq:combinedclusteringdistance}) would not.  
To compute $d_{ij}$ we combine two channels: PoTs and Motion Boundary
Histogram~\cite{wang_ICCV_2013}. 

We found this scheme extracts CMPs that reliably show similar
foreground motion and form good candidates for spatial alignment (sec.~\ref{sec:cmpevaluation}).
Restricting the search of CMPs within a behavior cluster
prunes unsuitable candidates (\eg a tiger jumping and one rolling on the ground).
Using only the top 10 pairs according to (\ref{eq:cmpcandidates}) 
further reduces the search space, extracting a manageable set of 
CMPs (\eg $3,000$ CMPs in a dataset of $100$ tiger shots, 
where we have to align 300 pairs of intervals after the behavior discovery
stage, sec.~\ref{sec:resultsalignment}).
The alternative strategy of trying to align 
all possible pairs of subsequences in the input shots is instead 
quadratic in the number of input frames (\httilde$300$ million), and thus
computationally impractical.

\subsection{Homography-based sequence alignment}
\label{sec:homography}  
Traditionally, homographies are used to model the mapping between two 
still images, and are estimated from a set
of 2-D point correspondences~\cite{Hartley00}.
Instead, we estimate the homography from
trajectory correspondences between two sequences (in a CMP).
We first review the traditional approach (sec.~\ref{sec:homographyimages}), and then present 
our extensions (sec.~\ref{sec:homographyvideos}-\ref{sec:homographyregularized}).

\subsubsection{Homography between still images}
\label{sec:homographyimages}
A 2-D homography $H_{uv}$ is a $3\times3$ matrix that
can be determined from four or 
more point correspondences $X_{u} \leftrightarrow X_{v}$ by solving
\begin{equation} X_u = H_{uv}X_v
\label{eq:fittinghomography}
\end{equation}
RANSAC~\cite{Fischler81} estimates a homography from a set
of putative correspondences $\mathcal{P}_{uv} = \{(x_u, y_u) \leftrightarrow
(x_v, y_v)\}$ that may include outliers. Traditionally, $\mathcal{P}_{uv}$
contains matches between local appearance descriptors, like SIFT~\cite{lowe04ijcv}.
RANSAC operates by running a large number of trials, each consisting of randomly sampling four point correspondences from $\mathcal{P}_{uv}$, fitting a homography to them, and counting the number of inliers it has in the whole set $\mathcal{P}_{uv}$. In the end, RANSAC returns the homography with the largest number of inliers.

\subsubsection{Homography between video sequences}
\label{sec:homographyvideos}  
\begin{figure}
\begin{center}
\includegraphics[scale = 2.5]{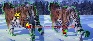}
\end{center}
 \caption{\small{
Modifying the TS descriptor. The TS descriptor is the concatenation
of the 2-D displacement vectors (green) of a trajectory across consecutive frames.
TS works well when aggregated in unordered representations
like Bag-of-Words~\cite{wang_ICCV_2013}, but matches found between
individual trajectories are not very robust, \eg
the TS descriptors for the trajectories on the torso of a tiger walking are almost identical.
We make TS more discriminative by appending the vector (yellow) between the trajectory
and the center of mass of the foreground mask (green) in the frame where
the trajectory starts (sec.~\ref{sec:homographyvideos}). 
We normalize this vector by the diagonal of the bounding box of the foreground mask
to preserve scale invariance.
 }}
\label{fig:augmentedTS}
\end{figure}

In video sequences, we use point trajectories as units for matching, instead of points in individual frames
(fig.~\ref{fig:sequencealignment} b).
We extract trajectories in each sequence and match them using a modified Trajectory Shape (TS) descriptor~\cite{wang_ICCV_2013} (fig.~\ref{fig:augmentedTS}). 
We match each trajectory in the first sequence to its nearest neighbor 
in the second with respect to Euclidean distance. 
We use trajectories which are $T=10$ frames long, and only match those starting in the same frame in both sequences.
Each trajectory match provides $T$ point correspondences (one per frame).

We consider two alternative ways to fit a homography to the trajectory matches,
called `Independent Matching' (IM) and `Temporal Matching' (TM).
IM treats the point correspondences generated by a single trajectory match independently during RANSAC. 
TM instead samples four {\em trajectory} matches at each RANSAC iteration,
and solves~\eqref{eq:fittinghomography} in the least squares sense using the $4\cdot T$
point correspondences.
A trajectory match is considered an outlier 
only if more than half of its point correspondences are outliers.
TM encourages geometric consistency over the duration of the CMP,
while IM could potentially overfit to point correspondences in just a few
frames. In practice, our experiments show that TM is superior to IM (sec.~\ref{sec:alignmentevaluation}).

We also considered matching PoTs across the sequences instead of individual trajectories,
but this is less efficient because each trajectory can be part of
many PoTs (we can build $O(n^2)$ PoTs out of $n$ trajectories).
Computationally, matching two sets of trajectories
of size $n$ and $m$ is $O(nm)$, while with PoTs it would be $O(n^2m^2)$ 

\subsubsection{Using the foreground mask as a regularizer}
\label{sec:homographyregularized}
\begin{figure}
\begin{center}
\includegraphics[scale = 0.25]{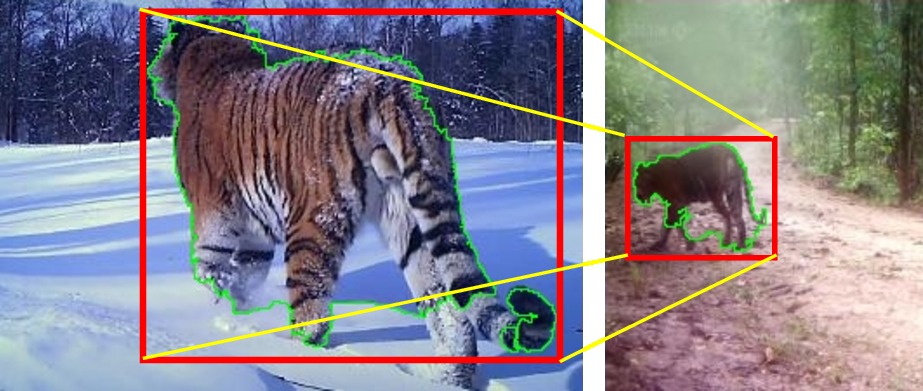}
\end{center}
 \caption{\small{
     Matching the corners between the bounding boxes of the foreground mask provides
     additional point correspondences (sec.~\ref{sec:homographyregularized}). These are too coarse to provide 
     a detailed spatial alignment between the two sequences and are also sensitive 
     to errors in the foreground masks, but they are useful when combined with 
     point correspondences from trajectory matches (fig.~\ref{fig:fgregularizer}).
 }}
\label{fig:fgmapping}
\end{figure}

\begin{figure}
\begin{center}
\includegraphics[scale = 0.3]{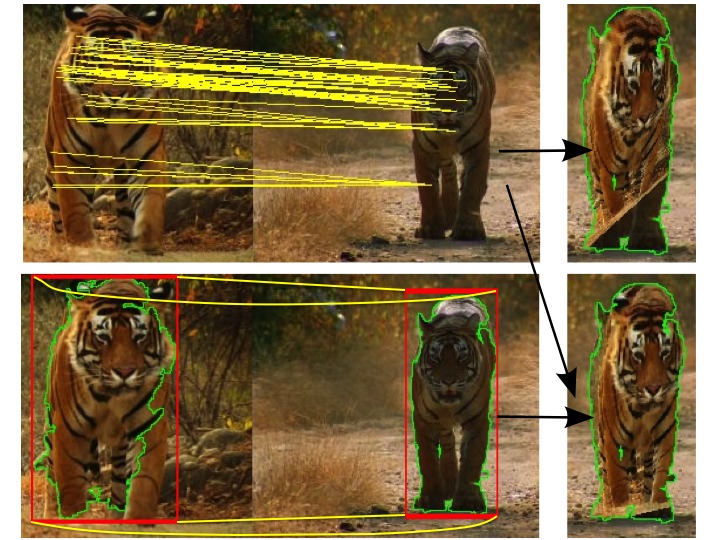}
\end{center}
 \caption{\small{
Top: Trajectory matches (yellow) often cover 
only part of the object. 
Here, the homography overfits to the correspondences
on the head, providing an incorrect mapping for the legs (right).
Bottom: Adding correspondences from the bounding boxes of the foreground masks~\cite{papazoglou13iccv} provides a more stable mapping (right, sec.~\ref{sec:homographyregularized}).
Note how also these correspondences are found automatically by our method (no manual intervention needed).
 }}
\label{fig:fgregularizer}
\end{figure}

The estimated homography tends to be inaccurate when the input
matches do not cover the entire foreground (fig.~\ref{fig:fgregularizer}).
To address this issue, we note that the bounding boxes of the foreground masks~\cite{papazoglou13iccv} induce
a very coarse global mapping (fig.~\ref{fig:fgmapping}).
Specifically, we include the correspondences between the bounding box
corners $F_{u} \leftrightarrow F_{v}$ in ~\eqref{eq:fittinghomography}:
\begin{equation}
\min_{H_{uv}} \|H_{uv}X_{v}- X_{u}\| + \|H_{uv}F_{v}- F_{u}\|~~~.
\label{eq:fittinghomomodified}
\end{equation}
This form of regularization makes our method much more stable (fig.~\ref{fig:fgregularizer}).

\subsection{Temporal TPS for sequence alignment}
\label{sec:tps}

We now present our approach to sequence alignment based on time-varying thin plate splines (TTPS). Unlike 
a homography, TTPS allows for local warping, which is more suitable for putting different object instances in correspondence.
We build on the popular TPS Robust Point Matching algorithm (TPS-RPM)~\cite{Chui03}, originally developed to align  point sets between two still images (sec.~\ref{sec:tpsrpm}).
We extend TPS-RPM to align two sequences of frames with a TPS that evolves smoothly over time (sec.~\ref{sec:tpssequence}).

\subsubsection{TPS-RPM}
\label{sec:tpsrpm}
A TPS $\mathrm{f}$ comprises an affine transformation $d$ and a non-rigid warp $w$. 
The mapping is a single closed-form function for the entire space,
with a smoothness term $\mathrm{L}(\mathrm{f})$ defined as the sum of the squares of the second
derivatives of $\mathrm{f}$ over the space~\cite{Chui03}.
Given two sets of points $\mathcal{U} = \{u_i\}$ and $\mathcal{V} = \{v_i\}$ in correspondence,
$\mathrm{f}$ can be estimated by minimizing
\begin{equation}
\label{eq:basictps}
\mathrm{E}(\mathrm{f}) = \sum_{i} || u_{i} - \mathrm{f}(v_{i})||^{2} + \lambda||\mathrm{L}(\mathrm{f})||.
\end{equation}
$\mathcal{U}$ and $\mathcal{V}$ are typically the position of detected image features (we use edge points, sec.~\ref{sec:tpssequence}).

As the point correspondences are typically not known beforehand, TPS-RPM jointly estimates $\mathrm{f}$
and a soft-assign correspondence matrix $M=\{m_{ij}\}$ 
by minimizing
\begin{equation}
\label{eq:tps-rpm}
\mathrm{E}(M, \mathrm{f}) = \sum_{i} \sum_{j} m_{ij}|| u_{i} - \mathrm{f}(v_{j})||^{2} + \lambda||\mathrm{L}(\mathrm{f})||.
\end{equation}
TPS-RPM alternates between updating $\mathrm{f}$ by keeping
$M$ fixed, and the converse. $M$ is continuous-valued, allowing
the algorithm to evolve through a continuous correspondence
space, rather than jumping around in the space
of binary matrices (hard correspondence).
It is updated by setting
$m_{ij}$ as a function of the distance between $u_{i}$
and $\mathrm{f}(v_{j})$~\cite{Chui03}. 
The TPS is updated by fitting $\mathrm{f}$ between
$\mathcal{V}$ and the current estimates $\mathcal{Y}$ of the corresponding points,
computed from $\mathcal{U}$ and $M$.

TPS-RPM optimizes \eqref{eq:tps-rpm} in a deterministic annealing \linebreak framework,
which enables finding a good solution even when starting from a relatively poor initialization.
The method is also robust to outliers in $\mathcal{U}$ and $\mathcal{V}$~\cite{Chui03}.

\subsubsection{Temporal TPS}
\label{sec:tpssequence}

\begin{figure}
\begin{center}
\includegraphics[scale =0.32]{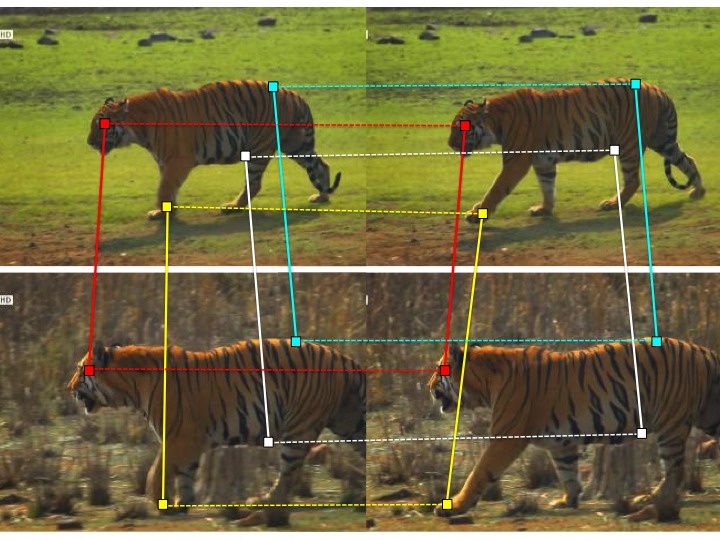} 
\end{center}
 \caption{\small{
Edge propagation using optical flow. In each sequence, we propagate
edge points extracted at time $t$ using optical flow,
independently in each sequence (dashed lines).
Our TTPS model (sec.~\ref{sec:tpssequence}) enforces
that the correspondences between edge points at time $t$ (solid lines)
be consistent with their propagated version at time $t+1$.
}
}
\label{fig:flowpropagation} \end{figure}

Our goal is to find a series of TPS
mappings $\mathcal{F}=\{\mathrm{f}^{1},\ldots,\mathrm{f}^{T} \}$,
one at each frame in the input sequences. We enforce temporal smoothness by
constraining each mapping to use a set of point correspondences consistent over time. 
Let $\mathcal{U}^t = \{u_i^t\}$ be the set of points for 
frame $t$ in the first sequence (with $\mathcal{V}^t$ defined analogously for the second sequence).
$\mathcal{U}^t$
contains both edge points extracted in
$t$ as well as edge points extracted in other frames and
propagated to $t$ via optical flow (fig.~\ref{fig:flowpropagation}). Each
$\mathcal{U}^t$ stores points in the same order, so that $u_i^{1}$
and  $u_i^{\tau}\forall \tau > 1$ are related by flow propagation
\footnote{ Consider a simple example with $T=2$, where we extract $10$ points
at $t=1$ and $20$ at $t=2$: $\mathcal{U}^1$ and
$\mathcal{U}^2$ contain $30$ points; the first $10$ in $\mathcal{U}^1$
are the point extracted at $t=1$, the next $20$ those extracted
at $t=2$ and propagated to $t=1$ with the backward flow; the first $10$ in $\mathcal{U}^2$
are the points extracted at $t=1$  propagated to $t=2$ with forward flow, the next $20$ those 
extracted at $t=2$.}.
%so that $u_i^t$ and
%$u_i^{\tau}$ are related by flow propagation.
We solve for the time-varying TPS $\mathcal{F}$ by minimizing
\begin{equation}
\label{eq:tpsflow}
\mathrm{E}(\mathcal{M}, \mathcal{F}) = \sum_{t}\! \left(\sum_{i}\! \sum_{j} m_{ij}^t|| u_{i}^{t} - \mathrm{f}^{t}(v_{j}^{t})||^{2} + \lambda||\mathrm{L}(\mathrm{f}^{t})|| \!\right) .
\end{equation}
subject to the constraint that $m_{ij}^1 = m_{ij}^{\tau}$ $\forall i,j,\tau > 1$. 
That is, if two points are in correspondence in frame $t$, they must still be in correspondence 
after being propagated to frame $\tau$.

\paragraph{Inference.}
Minimizing (\ref{eq:tpsflow}) is very challenging. In practice, we find an
approximate solution by first using TPS-RPM to fit a TPS $\mathrm{f}^\tau$ to
the edge points extracted at time $\tau$ only. This is initialized with the homography
found in sec.~\ref{sec:homographyregularized}.
Given the constraints on the $m_{ij}^t$,
$\mathrm{f}^{\tau}$ fixes the correspondences between $\mathcal{U}^t$ and $\mathcal{V}^t$
in all other frames.
We then fit the $\mathrm{f}^{t}$ $\forall t\neq \tau$ to these correspondences.
We repeat this process starting in each frame (\ie we try all $\tau \in [1,..,T]$), generating a total of $T$ TTPS candidates.
Finally, we return the one with the lowest energy (\ref{eq:tpsflow}).
Thanks to this efficient approximate inference, we can apply TTPS to align
thousands of CMPs.

\begin{figure}[t]
\begin{center}
\setlength{\tabcolsep}{0.8pt}
\begin{tabular}{ccccc}
\footnotesize{\textbf{fg mask}} & \footnotesize{\textbf{all edges}} & \footnotesize{\textbf{fg edges}} & \footnotesize{\textbf{edges*DT}} \\
\includegraphics[scale =0.3]{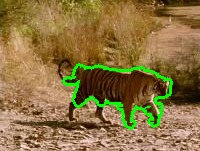}  &
\includegraphics[scale =0.3]{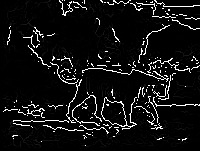}  &
\includegraphics[scale =0.3]{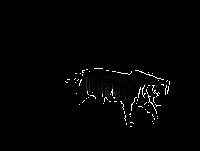}  &
\includegraphics[scale =0.3]{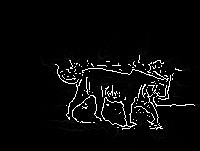}  \\
\small{(a)} & \small{(b)} & \small{(c)} & \small{(d)} \\
\end{tabular}
\end{center}
 \caption{\small{
Edge extraction (sec.~\ref{sec:tpssequence}). Using edges extracted from the entire image
confuses the TTPS fitting due to background edge points (b).
Using only edges on the foreground mask (c) loses 
useful edge points if the mask is inaccurate, \eg the missing legs in (a).
We instead weigh the edge strength (b) by the Distance
Transform (DT) with respect to the foreground mask. This is robust to errors
in the mask, while pruning most background edges (d).
}
}
\label{fig:tpsedges} \end{figure}

\paragraph{Foreground edge points.}
We extract edges using the edge detector~\cite{dollar13iccv} trained on the Berkeley Segmentation Dataset and Benchmark~\cite{martin01iccv}.
We remove clutter edges far from the object by multiplying the edge strength of each point with the Distance Transform (DT) of the image with respect to the foreground mask (\ie, the distance of each pixel to the closest point on the mask). We prune points scoring $\leq 0.2$.
This removes most background edges, and is robust to cases where the mask does not cover the complete object (fig~\ref{fig:tpsedges}).
To accelerate the TTPS fitting process, we further subsample the edge points to at most 1000 per frame.

\section{Related work}
\label{sec:related}

\subsection{Learning from videos}
\label{sec:relatedlearning}
A few recent works exploit video as a source of training data for object class detectors~\cite{Leistner11,prest12cvpr,Tang2013}.
They separate object instances from their background based on motion, thus reducing the need for manual bounding-box annotation. However, their use of video stops at segmentation.
They make no attempt at modeling articulated motion or finding common motion patterns across videos.
Ramanan~\etal~\cite{ramanan06pami} build a 2-D part-based model of an animal from one video. The model is a pictorial structure based on a 2-D kinematic chain of coarse rectangular segments.
Their method operates strictly on individual videos and therefore cannot learn class models. 
It is tested on just three simple videos containing only the animal from a single constant viewpoint.

In the domain of action recognition, classification is typically
formulated as a supervised problem~\cite{KTH,HMDB51,UCF101}.
Work on unsupervised motion analysis has largely
been restricted to the problem of dynamic scene 
analysis~\cite{kuettel10cvpr,HospedalesICCV09,Mahadevan2010,Wang2009PAMI,Hu2006PAMI,Zhao2011}.  
These works typically consider a fixed scene observed at a distance from a static
camera; the goal is to model the behavior of agents (typically pedestrians
and vehicles) and to detect anomalous events. Features typically consist of
optical flow at each pixel~\cite{HospedalesICCV09,kuettel10cvpr,Wang2009PAMI}
or single trajectories corresponding to tracked
objects~\cite{Hu2006PAMI,Zhao2011}.

Although many approaches do not easily transfer from the supervised to the
unsupervised setting, a breakthrough
from the action recognition literature that does is the concept of {\em dense trajectories}.
The idea of generating trajectories for each object from large numbers of KLT interest
points in order to model its articulation was simultaneously proposed by~\cite{Matikainen2009} and~\cite{Messing2009} for action recognition.
These ideas were extended and refined in the work on
tracklets~\cite{raptis10eccv} and DTFs~\cite{Wang_2011_CVPR}. Improved DTF (IDTFs)~\cite{wang_ICCV_2013} currently
provide state-of-the-art performance on video action recognition~\cite{THUMOS2014}.

\subsection{Representations related to PoTs}
In contrast to PoTs, most trajectory-based representations treat each trajectory in 
isolation~\cite{Wang_2011_CVPR,wang_ICCV_2013,Messing2009,Matikainen2009,raptis10eccv}.
Two exceptions are~\cite{Jiang2012,narayan14cvpr}.
Jiang \etal~\cite{Jiang2012} assign individual trajectories to a single codeword from a predefined
codebook (as in DTF works~\cite{Wang_2011_CVPR,wang_ICCV_2013}).
However, the codewords from a pair of trajectories are combined
into a `codeword pair' augmented by coarse information about the relative motion
and average location of the two trajectories. Yet, this pairwise analysis is
cursory: the selection of codewords is unchanged from the single-trajectory
case, and the descriptor thus lacks the fine-grained information about the
relative motion of the trajectories that PoTs provide.
Narayan \etal~\cite{narayan14cvpr} model Granger causality between trajectory codewords. 
Their global descriptor only captures pairwise statistics of codewords 
over a fixed-length temporal interval. In contrast, a PoT groups two trajectories into a single 
local feature, with a descriptor encoding their spatiotemporal arrangement. Hence, PoTs can be used to 
find point correspondences between different videos (fig.~\ref{fig:qualitative}).

The few remaining methods that propose pairwise representations 
employ them in a very different context.
Matikainen \etal~\cite{Matikainen2010} use spatial and temporal features
computed over pairs of sparse KLT trajectories to construct a two-level
codebook for action classification.
Dynamic-poselets~\cite{Wang2014} requires detailed manual annotations of human
skeletal structure on training data 
to define a descriptor for pairs of connected joints. 
Raptis \etal~\cite{raptis12cvpr} consider pairwise interactions between
clusters of trajectories, but their method also
requires detailed manual annotation for each action.
None of these approaches is suitable for 
unsupervised articulated motion discovery.
If we consider pairwise representations in still images,
Leordeanu~\etal~\cite{Leordeanu2007} learned object classes by matching pairs of contour points from one image to pairs in another.
Yang~\etal~\cite{Yang2010} computed statistics between local feature pairs
for food recognition, again in still images.

\subsection{Unsupervised behavior discovery}
\label{sec:relateddiscovery}
To our knowledge, only Yang \etal~\cite{Yang_PAMI_2013} considered the task of
unsupervised behavior discovery, albeit from manually trimmed videos.
Their method models human actions in terms of motion primitives discovered by 
clustering localized optical flow vectors, normalized with respect to the dominant translation of the object.
In contrast, PoTs capture the complex relationships between the motion of two different object parts. Furthermore, we describe motion at a more informative temporal scale by using
multi-frame trajectories instead of two-frame optical flow.
We compare experimentally to~\cite{Yang_PAMI_2013} on the KTH dataset~\cite{KTH} in sec.~\ref{sec:potsevaluation}.

\subsection{Spatial and temporal alignment}
\label{sec:relatedalignment}
Most works on spatial alignment focus on
aligning \emph{still images} for a variety of applications:
multi-view reconstruction~\cite{seitz2006comparison},
image stitching~\cite{Brown2007}, and object instance
recognition~\cite{ferrari06ijcv,lowe04ijcv}.
The traditional approach identifies candidate matches using a local appearance descriptor (\eg,
SIFT~\cite{lowe04ijcv}) with global geometric verification performed using \linebreak RANSAC~\cite{Fischler81,chum08pami}
or semi-local consistency checks~\cite{Schmid96,ferrari06ijcv,jegou08eccv}.
PatchMatch~\cite{barnes10eccv} and SIFT Flow~\cite{liu08eccv} generalize this notion to match patches between 
semantically similar scenes.

Our method differs from previous work on spatiotemporal \emph{video sequence
alignment}~\cite{caspi00cvpr,caspi06ijcv,ukrainitz06eccv} in several ways.
First, we find correspondences between different scenes, rather than between
different views of the same scene~\cite{caspi00cvpr,caspi06ijcv},
potentially at different times~\cite{evangelidis13pami}.
While the method in \cite{ukrainitz06eccv} is able to align actions across different scenes by directly maximizing local space-time correlations, it cannot handle the large intra-class appearance variations and diverse camera motions present in our videos.
As another key difference, all above approaches require temporally pre-segmented videos, \ie they
assume the two input videos show the same sequence of events in the same order and therefore can be aligned \emph{in their entirety}.
We instead operate with no available
temporal segmentation, which is why we assume that only small portions of the videos
can be aligned (the CMPs). Under stricter
assumptions, our method can potentially align much longer sequences.
%Instead, we operate on unsegmented videos and our method automatically finds which portions can be aligned.
Finally, these works have been evaluated only qualitatively
on 5-10 pairs of sequences, whereas we provide extensive 
quantitative analysis (sec.~\ref{sec:alignmentevaluation}).

Several approaches focus on finding the optimal temporal alignment (\ie frame-to-frame) between
two or more video sequences~\cite{tyutelaars04cvpr,wang14tog,douze15arxiv,dexter09bmvc,rao03iccv}.
Some of these works use a cost matrix to find the alignment
\cite{wang14tog,dexter09bmvc} similarly to our CMP candidate
extraction (sec.~\ref{sec:cmpcandidates}).
Also this class of methods
assumes that the input sequences can be aligned in their
entirety, or at least have a significant temporal overlap.

In the context of action recognition,
there has been work on matching spatiotemporal templates to
actor silhouettes~\cite{WeizmannActions,yilmaz05cvpr} or groupings
of supervoxels~\cite{ke07iccv}.
Our work is different because we map pixels from one unstructured video to another.
The method in~\cite{Jain2013} mines discriminative space-time patches and
matches them across videos.  It focuses on rough alignment using sparse
matches (typically one patch per clip), whereas we seek a finer, non-rigid spatial
alignment.
Other works on sequence alignment focus on temporal rather than spatial alignment~\cite{rao03iccv} or
target a very specific application, such as aligning presentation slides to videos of the corresponding 
lecture~\cite{fan11tip}.

A few methods use TPS for non-rigid point matching between still images~\cite{Chui03},
and to match shape models to images~\cite{ferrari10ijcv}. 
TPS were initially developed as a general purpose smooth functional mapping
for supervised learning~\cite{wahba90tps}.
The computer graphics community recently proposed \linebreak semi-automated
video morphing using TPS~\cite{liao14cgf}.
However, this method requires manual point correspondences as input, 
and it matches image brightness directly.

\section{Experiments}
\label{sec:experiments}

\subsection{Dataset}
\label{sec:dataset}

To evaluate our system, we assembled 
a new dataset of video shots for three highly 
articulated classes: tigers (500 shots), horses (100) and dogs (100).
The horse and dog shots are primarily low-resolution footage filmed by amateurs (YouTube), while
the tiger shots come from high-resolution National Geographic documentaries filmed by 
professionals.
This enables quantitative analysis on a large scale in two very different settings.

We automatically partition each tiger video into shots by thresholding color histogram 
differences in consecutive frames~\cite{kim09isce},
and kept only shots showing at least one tiger.
Horse and dog shots are sourced from the YouTube-Objects dataset~\cite{prest12cvpr},
where each shot contains at least one instance.

We provide two levels of ground-truth annotations: behavior labels to evaluate
PoTs (sec.~\ref{sec:potsevaluation}) and the behavior discovery 
stage (sec.~\ref{sec:discoveryevaluation}), and 2-D landmarks to evaluate the spatial
alignment stage (sec.~\ref{sec:alignmentevaluation}).
We publicly released this data at~\cite{delpero15cvpr-potswebpage}, where we also provide foreground masks for each shot
computed using~\cite{papazoglou13iccv}.

\paragraph{Behavior labels.}
We annotated all the frames in the dataset ($110,000$) with
the behavior displayed by the animal,
choosing from the labels in Table~\ref{table:resultspartitioning}.
As animals move over time, often a shot contains more than one label.
Therefore, we annotated each frame independently.
When a frame shows multiple behaviors, we chose the one that appears at the larger
scale (\eg, ``walk" over ``turn head", ``turn head" over ``blink").
If several animals are visible in the same frame, we annotated the behavior of the one closest to the camera.

\begin{figure}
\begin{center}
\includegraphics[scale = 0.25]{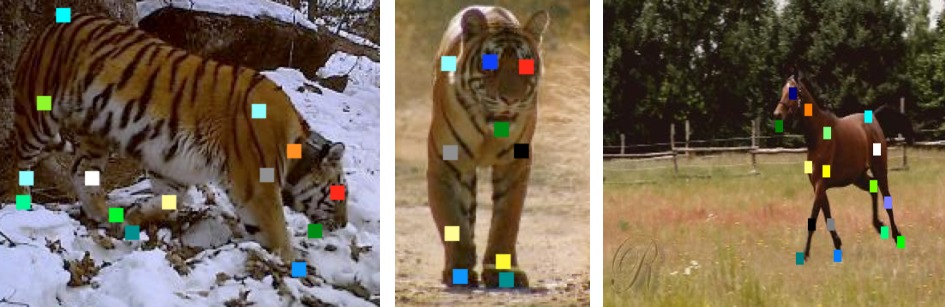}
\end{center}
 \caption{\small{
 Examples of annotated landmarks. A total of 19 points are marked when
 visible in over 17,000 frames for two different classes (horses and
 tigers). Our evaluation measure uses the landmarks to evaluate
 the quality of a sequence alignments (sec.~\ref{sec:alignmentevaluation}).  }}
\label{fig:landmarks}
\end{figure}

\paragraph{Landmarks.}
We annotated the 2-D location of 19 landmarks (fig.~\ref{fig:landmarks}) in 
all the $16,000$ frames of the horse class, 
and in $17,000$ of the tiger class (Tiger\_val, see below).
For horses we annotated: eyes (2), neck (1), chin (1), hooves (4), hips (4) and knees (4).
For tigers: eyes (2), neck (1), chin (1), ankles (4), feet (4) and knees (4).
We did not annotate occluded landmarks.
Unlike coarser annotations, such as bounding boxes,
landmarks enable evaluating the alignment of objects with
non-rigid parts with greater accuracy.
Again, if several animals are visible in the same frame, we annotated
the one closest to the camera.

\paragraph{Tiger subsets.}
We now define three different subsets of the tiger shots, which we
use throughout the experiments. \emph{Tiger\_all} denotes all tiger shots.
\emph{Tiger\_val} contains 100 randomly selected shots used to set the parameters of the methods we test. 
\emph{Tiger\_fg}  contains 100 manually selected shots in which the method of~\cite{papazoglou13iccv} produced accurate foreground masks (with no overlap with Tiger\_val).
We use Tiger\_fg to assess how sensitive the methods are to
the accuracy of the foreground masks.
All other subsets are instead representative of the average performance of~\cite{papazoglou13iccv}
(which is accurate on \httilde$55\%$ of the cases).

\subsection{Evaluation of PoTs}
\label{sec:potsevaluation}

We first evaluate PoTs (sec.~\ref{sec:pots}) in a simplified scenario
where we cluster intervals for which the correct single-behavior partitioning is given,
\ie, we partition shots at frames where the ground-truth behavior label changes. 
This allows us to evaluate the PoT representation separately 
from our method for automatic behavior discovery,
which does the partitioning automatically (sec.~\ref{sec:discoveryevaluation}).

\subsubsection{Evaluation protocol}
\label{sec:potsprotocol}
We compare PoTs to the state-of-the-art Improved Dense Trajectory Features (IDTFs)~\cite{wang_ICCV_2013}.
IDTFs combine four different feature
channels aligned with dense trajectories:
Trajectory shape (TS), Histogram of Oriented Gradients (HOG),
Histogram of Optical Flow (HOF), and Motion Boundary Histogram (MBH).
TS is the channel most related to PoTs, as it
encodes the displacement of an individual trajectory
across consecutive frames.
HOG is the only component based on appearance and not on motion.
We also compare against a version of IDTFs where only
trajectories on the foreground masks are used, which we
call fg-IDTFs. We use the same
point tracker~\cite{wang_ICCV_2013} to extract both
IDTFs and PoTs. For PoTs, we do not remove trajectories that are static or are caused by the motion
of the camera. Removing these trajectories improves the performances
of IDTFs~\cite{wang_ICCV_2013}, but in our case they are useful as potential anchors.

We adopt two criteria commonly used for evaluating
clustering methods: \emph{purity} and \emph{Adjusted Rand Index} (ARI)~\cite{rand71jasa}.
Purity is the number of items correctly clustered
divided by the total number of items
(an item is correctly clustered if its label coincides 
with the most frequent label in its cluster).
While purity is easy to interpret, it only penalizes
assigning two items with different labels to the same cluster.
The ARI instead also penalizes
putting two items with the same label in different clusters.
Further, it is adjusted such that a random clustering
will score close to $0$. 
It is considered a better way to evaluate clustering methods
by the statistics community~\cite{Lawrence_JOC_1985,santos09icann}.

\paragraph{Parameter setting.}
We use Tiger\_val to set the PoT selection threshold $\theta_{P}$
(sec.~\ref{sec:potselection}) and the PoT codebook
size $V$ (sec.~\ref{sec:timeclustering}) using coarse grid search.
As objective function, we used the ARI achieved by our 
method when the number of clusters is equal to the true number of behaviors.
We used interval $[0.05,0.35]$ with a step of $0.05$ for
$\theta_{P}$,  and $[800,8000]$ with a step
of $800$ for $V$. 
Grid search selects $\theta_{P}=0.15, V=800$ and we use
these values in all experiments on all classes.
In practice, performance is very similar for a wide range
of parameters: $0.1 \leq \theta_{P} \leq 0.25$ and $800 \leq V \leq 1600$.
We tuned the IDTFs codebook size analogously and found that $4000$ codewords work best.
Interestingly, the same value is chosen by Wang~\etal.~\cite{wang_ICCV_2013}
on completely different data.

\begin{figure*}[t]
\begin{center}
\setlength{\tabcolsep}{0.5pt}
\begin{tabular}{c c c c}
\small{(a)} & \small{(b)} & \small{(c)} & \small{(d)}\\
\includegraphics[scale =0.13]{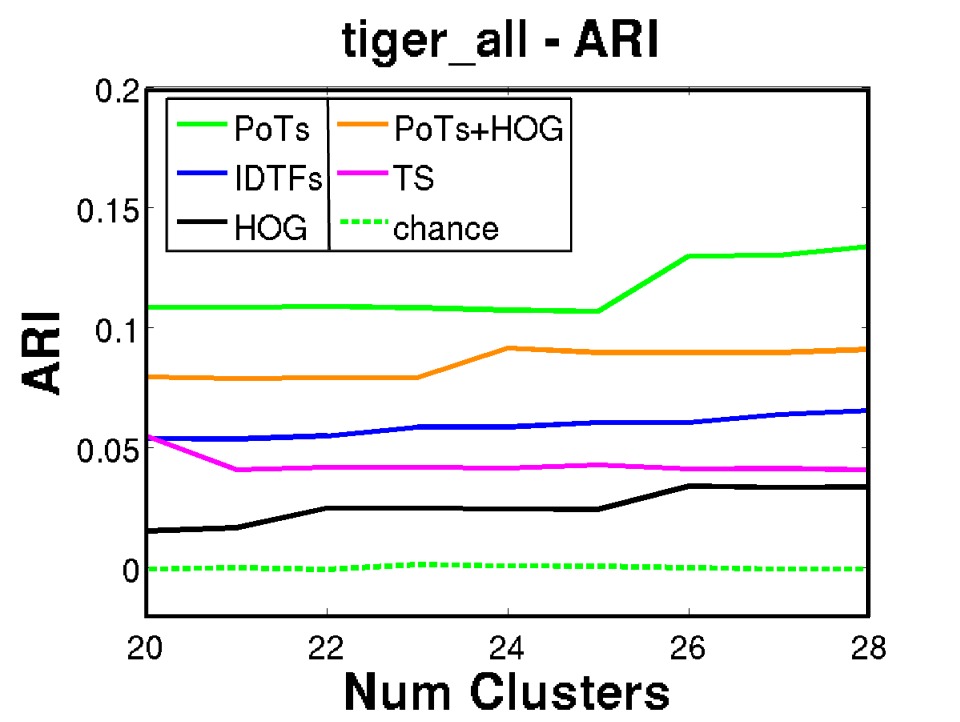}&
\includegraphics[scale =0.13]{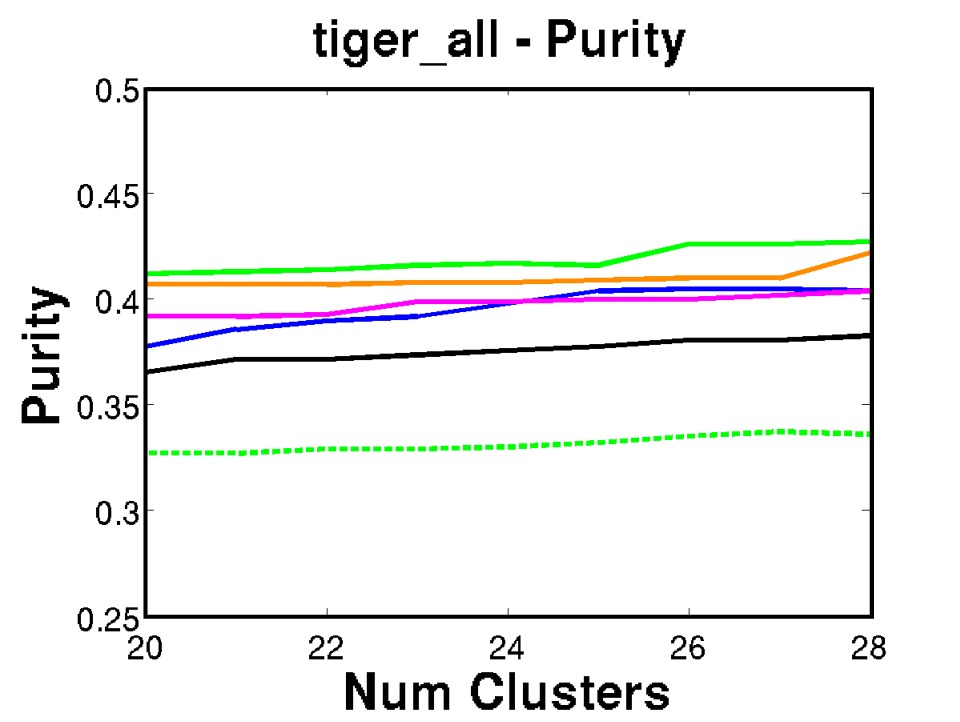}&
\includegraphics[scale =0.13]{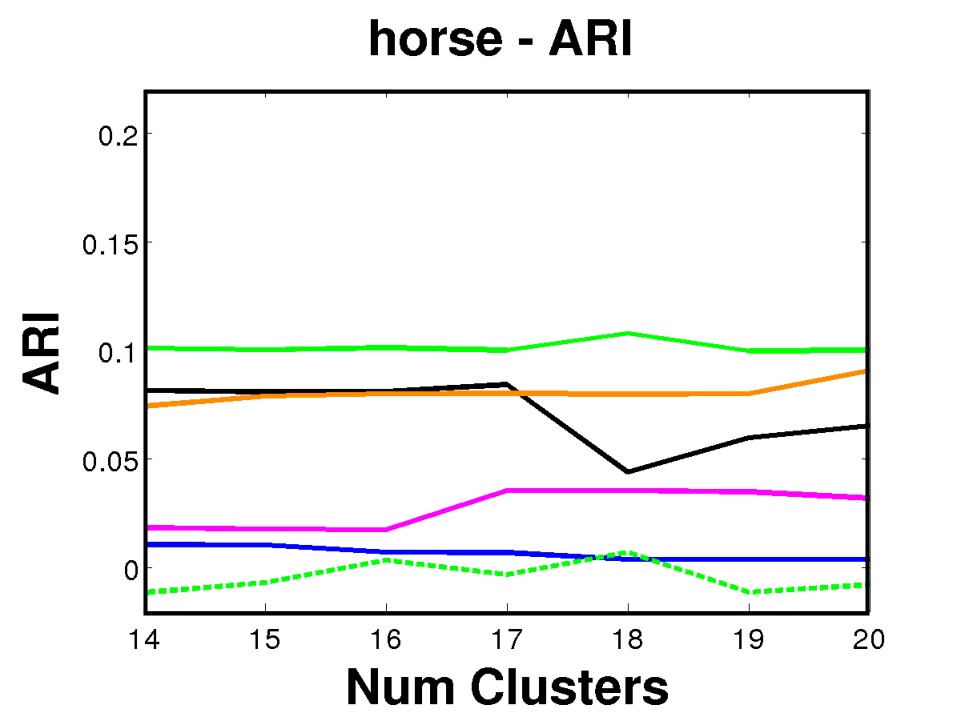}&
\includegraphics[scale =0.13]{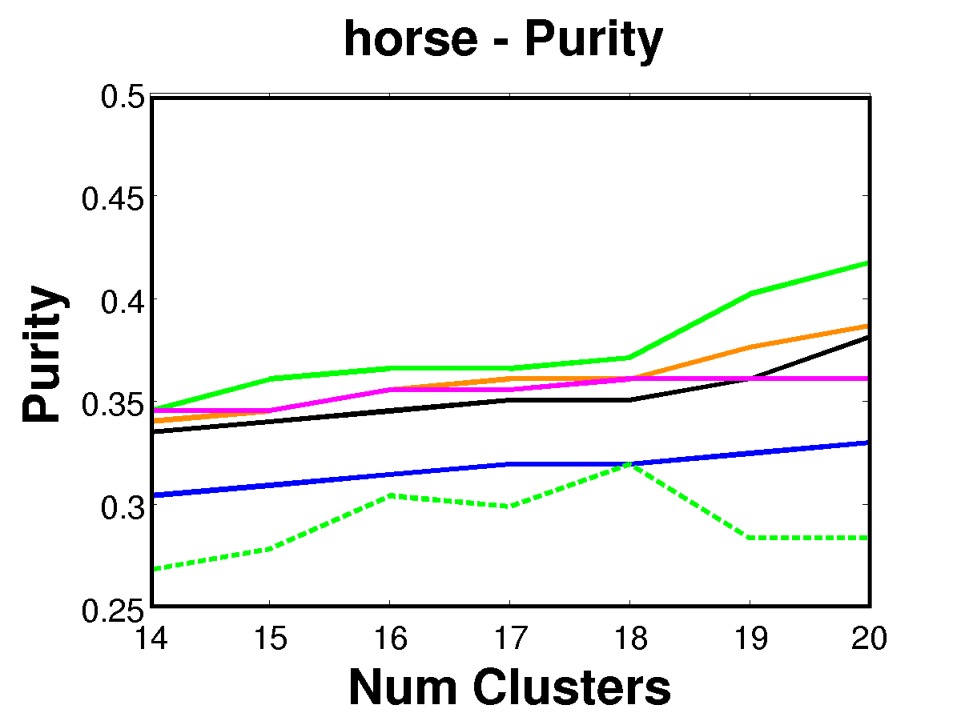}\\
\small{(e)} & \small{(f)} & \small{(g)} & \small{(h)} \\
\includegraphics[scale =0.13]{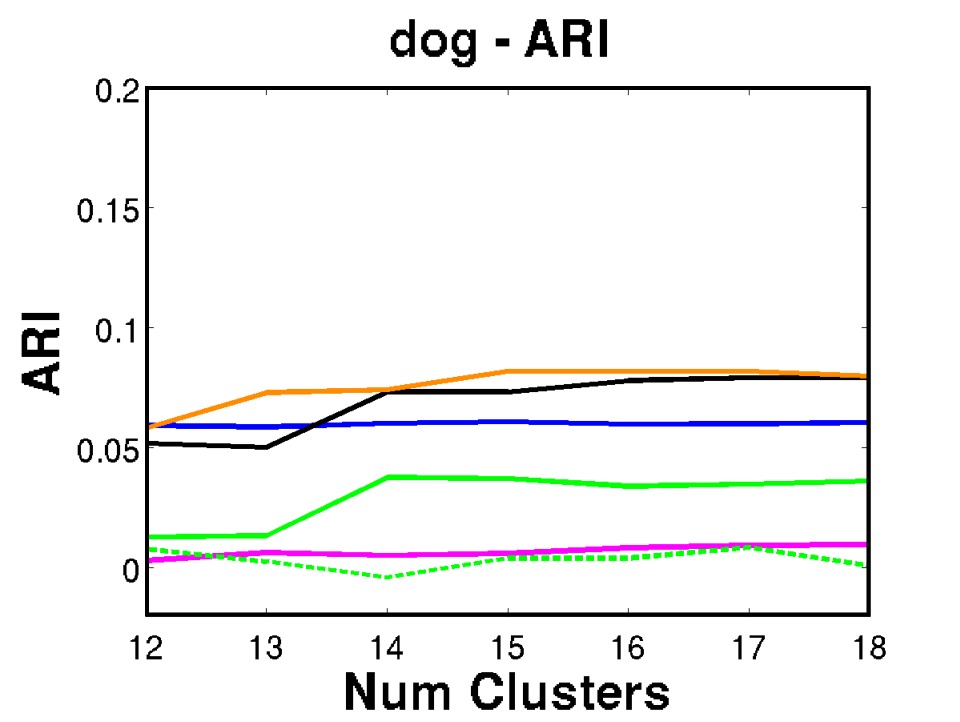}&
\includegraphics[scale =0.13]{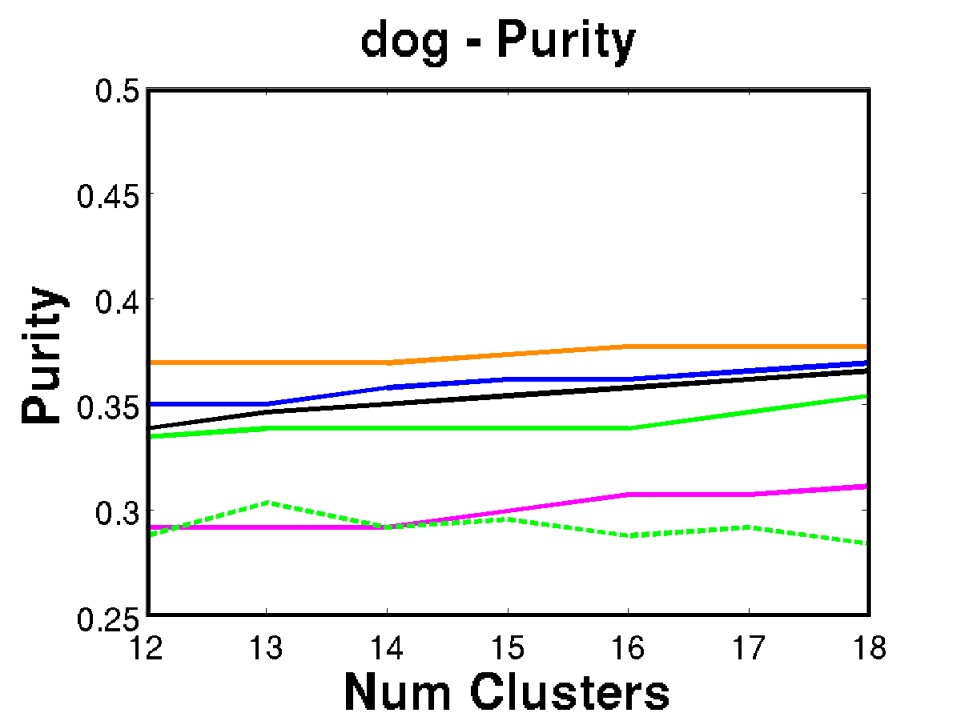} &
\includegraphics[scale =0.13]{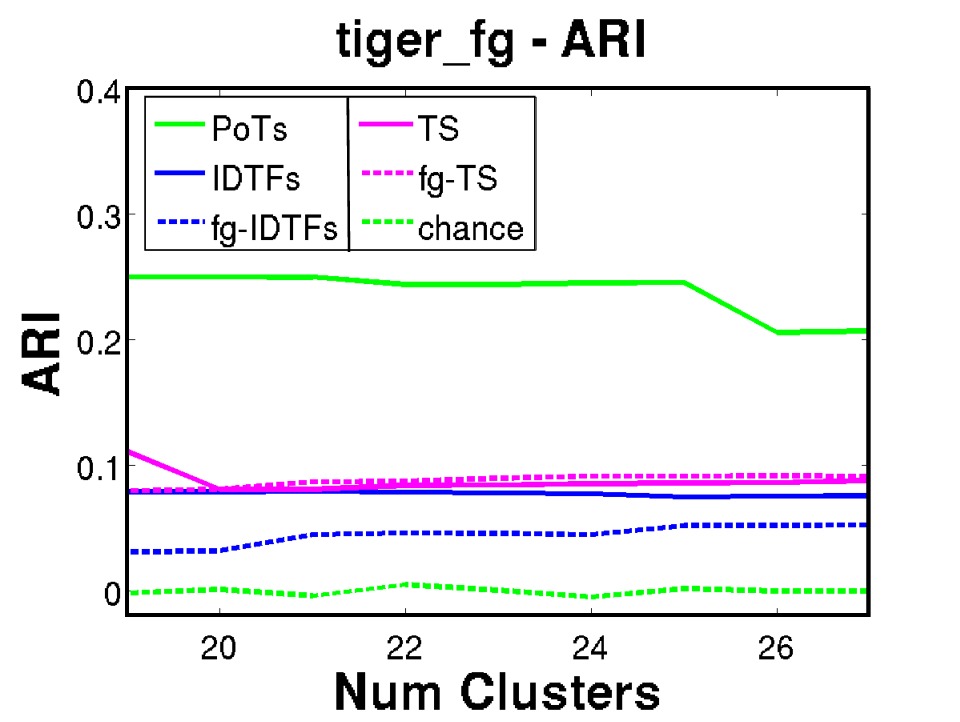} &
\includegraphics[scale =0.13]{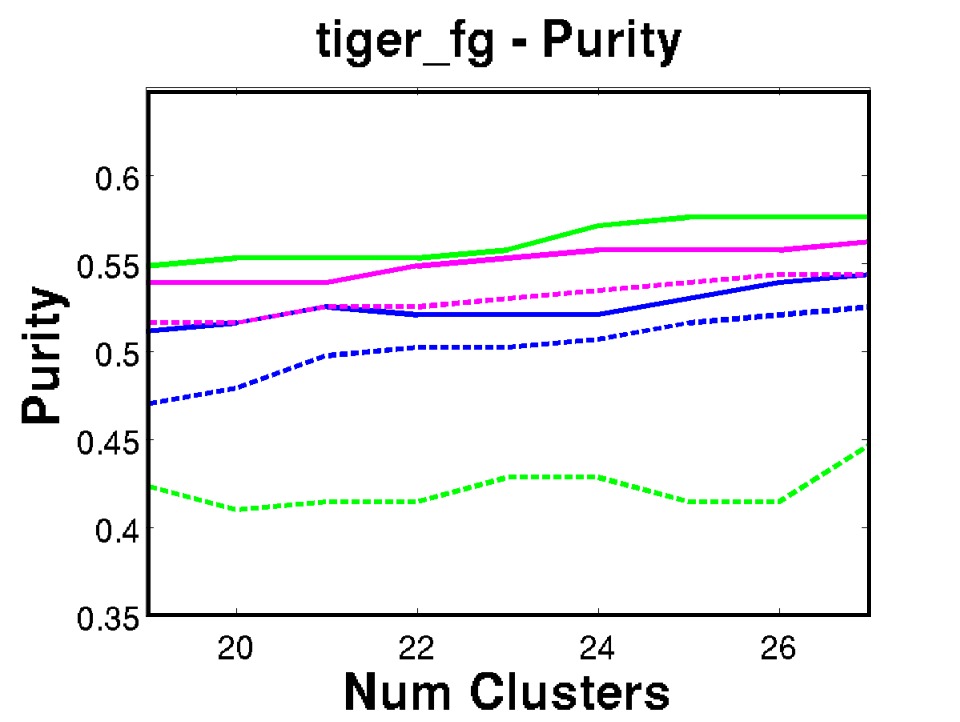} \\
\hline
\vspace{4pt}
\small{(i)} & \small{(j)} & \small{(k)} & \small{(l)} \\
\includegraphics[scale =0.13]{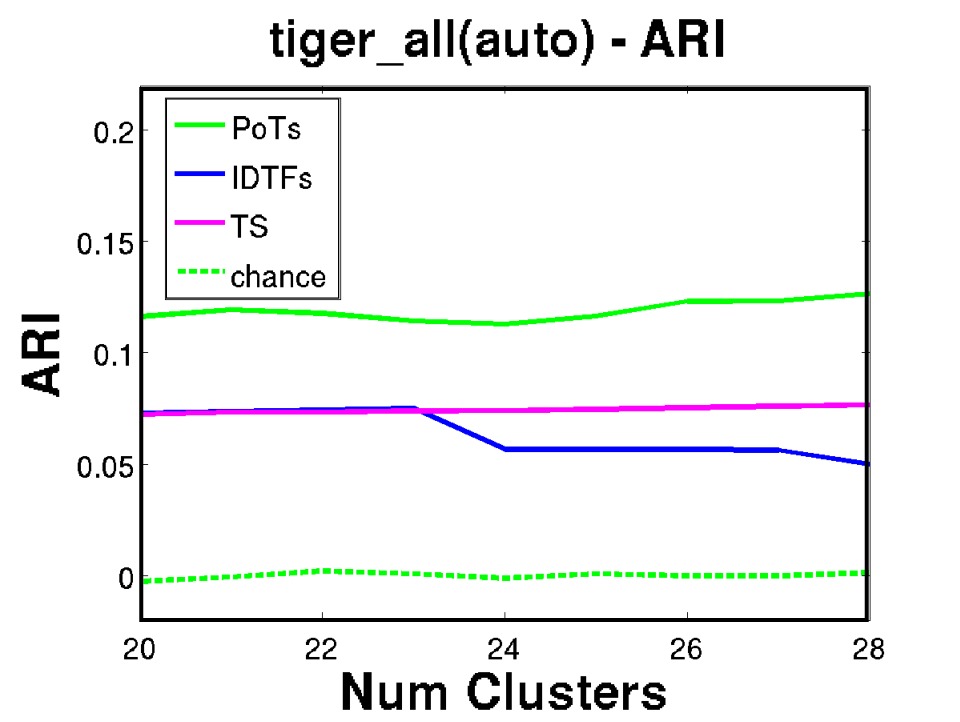}&
\includegraphics[scale =0.13]{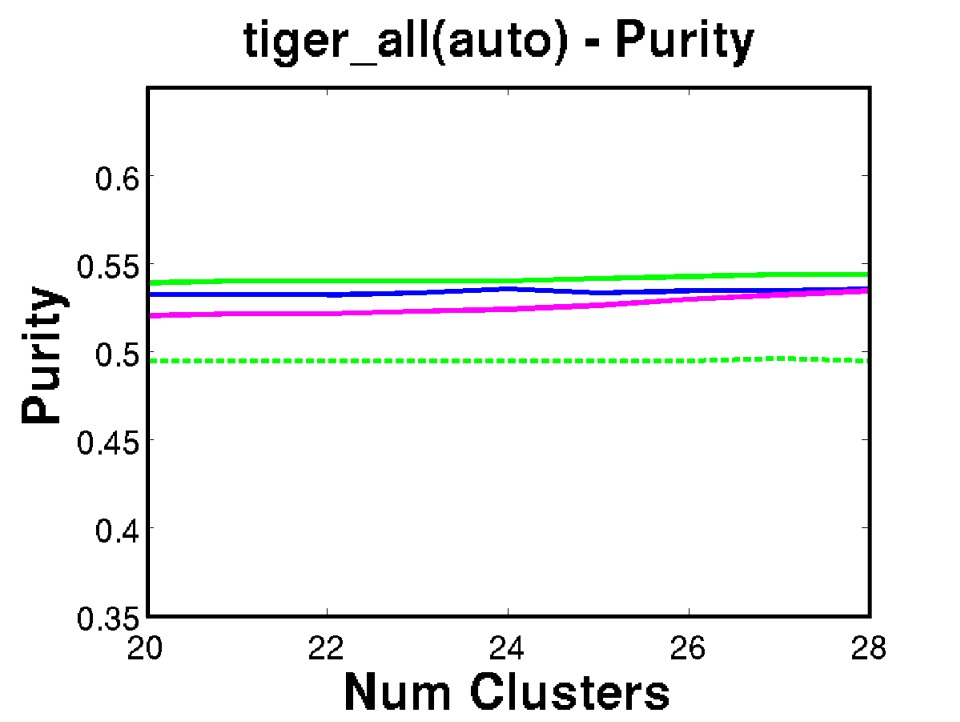}&
\includegraphics[scale =0.13]{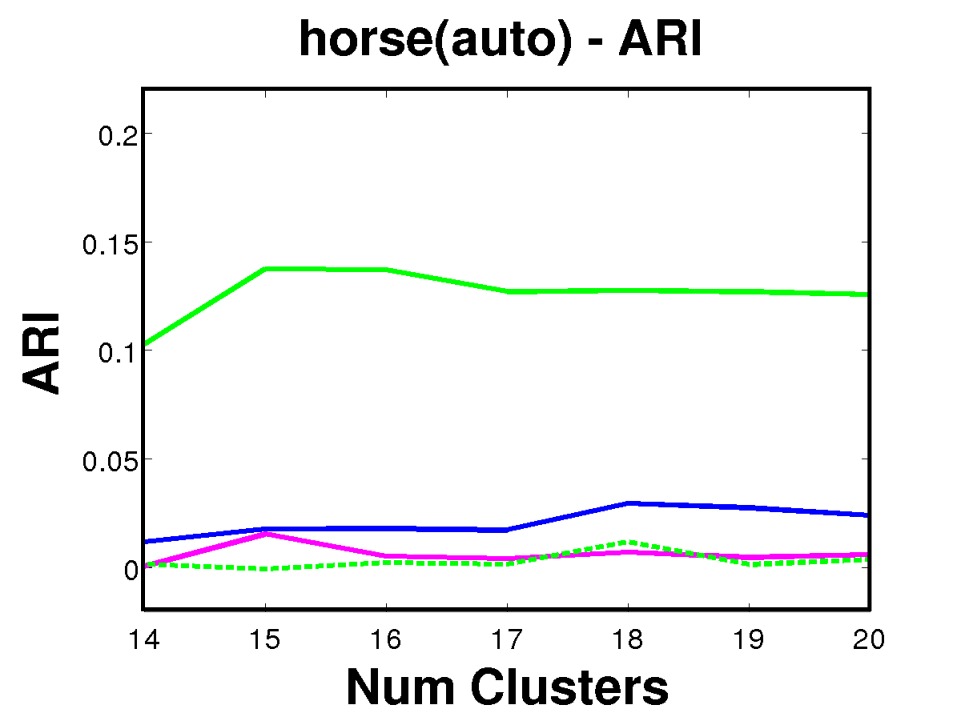}& 
\includegraphics[scale =0.13]{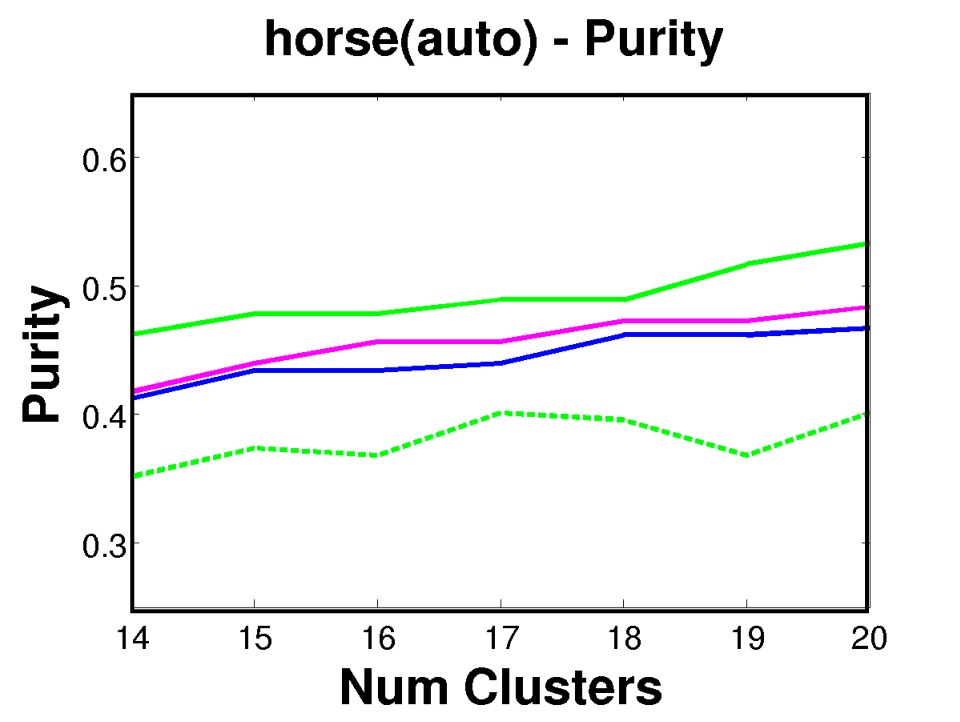}\\
\end{tabular}
\end{center}
 \caption{\small{
Results of clustering intervals using different descriptors (sec.~\ref{sec:potsresults}),
evaluated on Adjusted Rand Index (ARI) and purity (sec.~\ref{sec:potsprotocol}). 
PoTs result in better clusters than 
IDTFs~\cite{wang_ICCV_2013} on tigers and horses (a-d). 
Adding appearance (PoTs+HOG) is detrimental on these two classes, 
but improves performance on dogs (e-f).
IDTFs perform well for dogs, primarily due to the
contribution of the HOG channel: compare the full descriptor (blue), with
the HOG channel only (black) and the trajectory shape channel TS (magenta). For
all classes, PoTs+HOG performs better than IDTFs.
The gap between IDTFs and PoTs increases
on tiger\_fg, where we ensured the segmentation is accurate (g-h).
Here, PoTs also outperform IDTFs extracted on the foreground mask only (fg-IDTFS).
PoTs also generate higher-quality clusters than the other methods
when we cluster automatically partitioned intervals (i-l, sec~\ref{sec:discoveryevaluation}).
}
}
\label{fig:resultspots}
\end{figure*}

\subsubsection{Results}
\label{sec:potsresults}
We compare clustering using BoWs of PoTs to 
using BoWs of IDTFs in fig.~\ref{fig:resultspots}, (a-h).
As the true number of clusters
is usually not known a priori, each plot
shows performance as a function of the number of clusters. 
The mid value on the horizontal axis corresponds
to the true number of behaviors ($23$ for tigers, $17$ for horses, $15$ for dogs).

For tigers and horses, the clusters found using PoTs are better in both purity and ARI, compared to using IDTFs (fig.~\ref{fig:resultspots}, a-d).
Consider now the individual IDTFs channels.
On tigers, the HOG channel  performs poorly, and adding it to PoTs (PoTs+HOG) performs
worse than PoTs alone.
Appearance is in general not suitable for distinguishing between fine-grained behaviors.
It is particularly misleading when different object instances have similar color and texture
(like tigers).
The HOF and MBH channels of IDTF perform poorly on their own and are not shown in the plot.

The gain over IDTFs is larger on Tiger\_fg (g-h), where PoTs benefit from the accurate
foreground masks. Here, PoTs also outperform fg-IDTFs, showing
that the power of our representation resides in the principled
use of pairs of trajectories, not just in exploiting foreground masks to remove background trajectories.
Moreover, all other results (a-f) show that PoTs can also cope with imperfect masks.

\begin{table*}[t]
\begin{center}
\scriptsize
\setlength{\tabcolsep}{2.8pt}
\begin{tabular}{lccccccccccccccccccccccc}
\toprule
\textbf{tiger} &\textbf{walk} &\textbf{turn}  &\textbf{sit} &\textbf{tilt} &\textbf{stand} &\textbf{drag} &\textbf{wag} &\textbf{walk} &\textbf{run} &\textbf{turn} &\textbf{jump} &\textbf{raise} &\textbf{open} &\textbf{close} &\textbf{blink} &\textbf{slide} &\textbf{drink} &\textbf{chew} &\textbf{lick} &\textbf{climb} &\textbf{roll} &\textbf{scratch} &\textbf{swim} \tabularnewline
\textbf{partitions} &\textbf{} &\textbf{head}  &\textbf{down} &\textbf{head} &\textbf{up} &\textbf{} &\textbf{tail} &\textbf{back} &\textbf{} &\textbf{} &\textbf{} &\textbf{paw} &\textbf{mouth} &\textbf{mouth} &\textbf{} &\textbf{leg} &\textbf{} &\textbf{} &\textbf{} &\textbf{} &\textbf{} &\textbf{} &\textbf{} \tabularnewline
\midrule
\textbf{\tiny whole shots} & 259 & 24 & 5 & 17 & 5 & 4 & 1 & 5 & 16 & 3 & 9 & 0 & 4 & 0 & 0 & 1 & 7 & 4 & 6 & 1 & 3 & 1 & 3 \tabularnewline
\textbf{\tiny pauses} & 272 & 76 & 11 & 30 & 9 & 4 & 2 & 6 & 16 & 2 & 11 & 2 & 11 & 3 & 10 & 2 & 7 & 5 & 7 & 1 & 5 & 1 & 3 \tabularnewline
\textbf{\tiny pauses+periods} & \textbf{273} & \textbf{80} & \textbf{13} & \textbf{33} & \textbf{9} & \textbf{4} & \textbf{2} & \textbf{6} & \textbf{16} & \textbf{3} & \textbf{11} & \textbf{2} & \textbf{12} & \textbf{3} & \textbf{11} & \textbf{3} & \textbf{8} & \textbf{5} & \textbf{7} & \textbf{1} & \textbf{5} & \textbf{1} & \textbf{3} \tabularnewline
\textbf{\tiny ground truth} & 289 & 148 & 27 & 77 & 24 & 4 & 4 & 10 & 23 & 18 & 20 & 6 & 40 & 28 & 39 & 13 & 12 & 7 & 19 & 1 & 5 & 2 & 3 \tabularnewline
\bottomrule
\end{tabular}
\begin{tabular}{lccccccccccccccccccccccc}
\toprule
\textbf{horse} &\textbf{walk} &\textbf{turn}  &\textbf{sit} &\textbf{tilt} &\textbf{stand} &\textbf{drag} &\textbf{walk} &\textbf{gallop} &\textbf{turn} &\textbf{jump} &\textbf{raise} &\textbf{trot} &\textbf{piaffe} &\textbf{jump} &\textbf{graze} &\textbf{rodeo} &\textbf{rolling} \tabularnewline
\textbf{partitions} &\textbf{} &\textbf{head}  &\textbf{down} &\textbf{head} &\textbf{up} &\textbf{} &\textbf{back} &\textbf{} &\textbf{} &\textbf{} &\textbf{paw} &\textbf{} &\textbf{} &\textbf{hurdles} &\textbf{} &\textbf{} &\textbf{} \tabularnewline
\midrule
\textbf{\tiny whole shots} & 16 & 2 & 1 & 5 & 1 & 3 & 0 & 30 & 2 & 1 & 0 & 17 & 3 & 6 & 1 & 4 & 1 \tabularnewline
\textbf{\tiny pauses} & 16 & 2 & 1 & 7 & 1 & 3 & 0 & 31 & 3 & 1 & 0 & 20 & 3 & 6 & 1 & 4 & 1  \tabularnewline
\textbf{\tiny pauses+periods} & \textbf{19} & \textbf{4} & \textbf{1} & \textbf{8} & \textbf{2} & \textbf{3} & \textbf{1} & \textbf{33} & \textbf{4} & \textbf{1} & \textbf{0} & \textbf{20} & \textbf{3} & \textbf{6} & \textbf{1} & \textbf{4} & \textbf{1} & \tabularnewline
\textbf{\tiny ground truth} & 27 & 11 & 1 & 12 & 2 & 3 & 1 & 39 & 11 & 2 & 1 & 22 & 3 & 7 & 6 & 4 & 3 \tabularnewline
\bottomrule
\end{tabular}
\begin{tabular}{lcccccccccccccccc}
\toprule
\textbf{dog} &\textbf{walk} &\textbf{turn}  &\textbf{sit} &\textbf{tilt} &\textbf{stand} &\textbf{walk} &\textbf{run} &\textbf{turn} &\textbf{jump} &\textbf{open} &\textbf{close} &\textbf{blink} &\textbf{slide} &\textbf{lick} &\textbf{push} \tabularnewline
\textbf{partitions} &\textbf{} &\textbf{head}  &\textbf{down} &\textbf{head} &\textbf{up} &\textbf{back} &\textbf{} &\textbf{} &\textbf{} &\textbf{mouth} &\textbf{mouth} &\textbf{} &\textbf{leg} &\textbf{} &\textbf{skateboard} \tabularnewline
\midrule
\textbf{\tiny whole shots} & 25 & 4 & 0 & 1 & 0 & 0 & 10 & 0 & 4 & 0 & 0 & 0 & 0 & 0 & 13  \tabularnewline
\textbf{\tiny pauses} & 29 & 5 & 0 & 1 & 0 & 0 & 12 & 2 & 5 & 0  & 0 & 0  & 0 & 0 & 16  \tabularnewline
\textbf{\tiny pauses+periods} & \textbf{29} & \textbf{9} & \textbf{0}  & \textbf{3} & \textbf{0} & \textbf{1} & \textbf{13} & \textbf{5} & \textbf{5} & \textbf{0} & \textbf{0} & \textbf{0} & \textbf{0} & \textbf{0} & \textbf{16} \tabularnewline
\textbf{\tiny ground truth} & 39 & 25 & 1 & 12 & 1 & 2 & 20 & 14 & 8 & 2 & 1 & 2 & 1 & 1 & 19  \tabularnewline
\bottomrule
\end{tabular}
\end{center}
 \caption{\small
Number of intervals recovered per behavior on tigers (top), horses (middle) and dogs (bottom). Pauses+periodicity consistently dominates others (sec.~\ref{sec:discoveryevaluation}).
 \label{table:resultspartitioning}
}
\end{table*}

For the dog class, IDTFs perform better than PoTs
(fig.~\ref{fig:resultspots}, e-f). However,
HOG is doing most of the work in this case.
The dog shots come from only eight different videos, each showing
one particular dog performing $1$--$2$ behaviors in the same scene.
Hence, HOG performs well by trivially clustering together
intervals from the same video.
When we equip PoTs with HOG, they outperform the complete IDTFs.
Additionally, if we consider trajectory motion alone
PoTs outperform TS, further confirming that PoTs are a more suitable
representation for articulated motion.

Results on tigers and horses showed that adding appearance features can be detrimental,
since there is little correlation between a behavior
and the appearance of the animal and/or the background.
This is not the case for the dog class, where the shots come from only eight different scenes,  
compared to more than fifty for horses, and several hundreds for tigers.
However, it shows that PoTs and appearance features are complementary: 
when appearance should be beneficial, 
we see the expected performance boost by adding this additional
information. This is potentially useful for traditional
action recognition tasks~\cite{UCF101,Sports1M}, where
many activities strongly correlate with the background and the apparel involved
(\eg, diving can be recognized from the appearance of swimsuits, or a diving board
with a pool below). Last, we note that we use the same PoT parameters on all
datasets (set on Tiger\_val, sec.~\ref{sec:potsprotocol}), showing that our
representation generalizes across classes.

\paragraph{Comparison to motion primitives~\cite{Yang_PAMI_2013}.}
Last, we compare to the method of~\cite{Yang_PAMI_2013}, which is based on motion 
primitives (sec~\ref{sec:relateddiscovery}).
Since they did not release their method, we compare to the results they report on the KTH dataset~\cite{KTH} in their setting.
The KTH dataset contains $100$ shots for each of six different human actions (\eg walking, hand clapping).
As before, we cluster all shots using the PoT representation: for the true number 
of clusters (6), we achieve 59\% purity, compared to their 38\% (fig. 9 in~\cite{Yang_PAMI_2013}).
For this experiment, we incorporated an R-CNN person detector~\cite{girshick14cvpr}
into~\cite{papazoglou13iccv} to better segment the actors.

\begin{table}
 \vspace{-5pt}
\begin{center}
\small
\setlength{\tabcolsep}{1.8pt}
\begin{tabular}{lcccc}
\toprule
&\textbf{whole shots} & \textbf{pauses} & \textbf{pauses+periods} & \textbf{ground truth} \tabularnewline
\midrule
tiger \# intervals &480 & 719 & \textbf{885} & 1026 \tabularnewline
tiger uniformity &0.78 & 0.85 & \textbf{0.87} & 1 \tabularnewline
\hline
\midrule
horse \# intervals & 96 & 117 & \textbf{184} & 194 \tabularnewline
horse uniformity & 0.82 & 0.83 & \textbf{0.89} & 1 \tabularnewline
\hline
\midrule
dog \# intervals & 80 & 115 & \textbf{219} & 260 \tabularnewline
dog uniformity &0.72 & 0.80 & \textbf{0.88} & 1 \tabularnewline
\bottomrule
\end{tabular}
\end{center}
 \caption{\small
Interval uniformity for different partitioning methods.
Pauses+periods consistently outperforms alternatives (sec.~\ref{sec:discoveryevaluation}).
 \label{table:intervaluniformity}}
\end{table}

\subsection{Evaluation of behavior discovery}
\label{sec:discoveryevaluation}
We first evaluate our method for partitioning shots into single-behavior intervals 
(sec.~\ref{sec:timepartitioning}).
Let the {uniformity} of an interval be the number of frames with the
most frequent label in it, divided by the total number of frames.
The combination of pauses and periodicity partitioning improves the
baseline average interval uniformity of the original, unpartitioned shots
(Table~\ref{table:intervaluniformity}).
This is very promising, since the average uniformity is near $90\%$,
and the number of intervals found approaches the ground-truth number. 
In Table~\ref{table:resultspartitioning} we report the number of single-behavior 
intervals found by each method, grouped by behavior.
We only increase the count for intervals from different shots,
otherwise we could approach the ground-truth number by simply partitioning one continuous behavior 
into smaller and smaller pieces (\eg if our method returns three intervals
from the same shot whose ground-truth label is ``walking", we increase
the count for ``walking" in Table~\ref{table:intervaluniformity} only by one).
We chose this counting method because our ultimate goal is to find instances of the same behavior 
performed by different object instances.
Clustering whole shots would lose many behaviors, and only a few dominant ones such as walking would emerge.
Our method instead finds intervals for almost all behavior types.

\begin{figure*}[t]
\begin{center}
\includegraphics[scale =0.33]{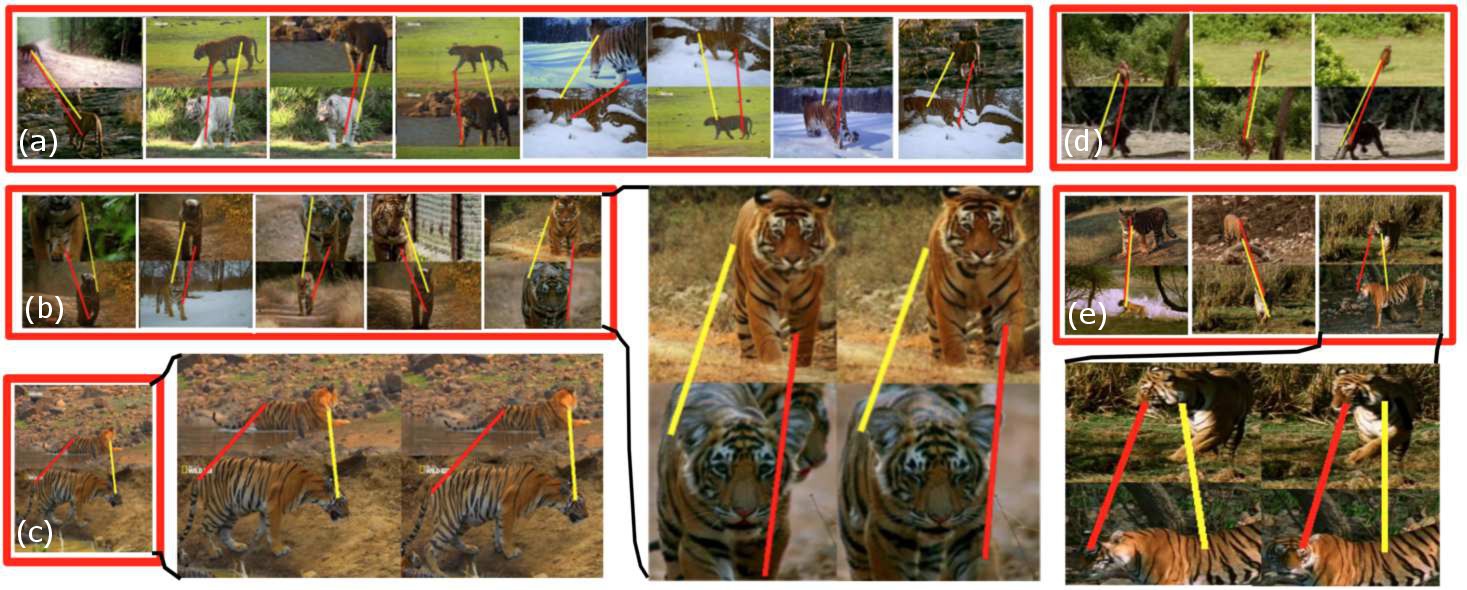}
\end{center}
 \caption{\small{
Behaviors discovered by clustering consistent motion patterns (sec.~\ref{sec:discoveryevaluation}).
Each red rectangle displays a few pairs of intervals
from one cluster, on which we connect the anchors (yellow)
and swings (red) of two individual PoTs that are close in descriptor space.
The enlarged version show how the connected PoTs evolve through time,
and give a snapshot of one representative motion pattern for each cluster.
%For example, the rotation of the head wrt other part
%of the body in the ``turning head" cluster" (bottom right).
The behaviors shown are: two different ways of walking (a,b), sitting down
(c), running (d), and turning head (e).
Video showing behavior clusters for all classes are available
on our website~\cite{delpero15cvpr-potswebpage}.
 }
 }
\label{fig:qualitative}
\end{figure*}

Last, we report purity and ARI for the clusters of partitioned intervals (fig.~\ref{fig:resultspots}, i-l). 
As ground-truth label for a partitioned interval, we use the ground-truth label of the majority of the frames in it.
PoTs outperform IDTFs on tigers and horses also in this setting.
To make this comparison fair, we evaluate IDTFs and PoTs after using
the same partitioning method (pauses+periodicity).
We show a few qualitative examples of the discovered behavior clusters in fig.~\ref{fig:qualitative}.

\subsection{Evaluation of sequence alignment}
\label{sec:alignmentevaluation}

The input of this experiment are the clusters of intervals
discovered by our method (step 3 in fig.~\ref{fig:architecture}).
We set the number of clusters to be a fourth of the number of intervals
in step 2. 
With this settings, the purity of the discovered clusters
is above $0.7$ (CMP extraction in step 4 benefits from having reasonably pure 
clusters as input). 
For the tiger class we only cluster the intervals in Tiger\_val, since
this is the only subset of the tiger class with landmark annotations (we
use all intervals for horses). 

We now introduce an alignment error measure
(sec.~\ref{sec:alignmenterror}), which we use to evaluate
CMP extraction (sec.~\ref{sec:cmpevaluation}) and alignment (sec.~\ref{sec:resultsalignment}).

\subsubsection{Alignment error}
\label{sec:alignmenterror}
We evaluate the mapping found between the two sequences in a CMP as follows.
For each frame, we map each landmark in the first sequence onto the second and compute the Euclidean distance
to its ground-truth location.
The error for the landmark is the average between this distance and the reverse 
(\ie, when we map the landmark from the second sequence into the first).
We normalize the error by the scale of the object,
defined as the maximum distance between any two landmarks in the frame.
The overall alignment error is the average error of all visible landmarks over all frames.

\begin{figure}
\begin{center}
\includegraphics[scale = 0.38]{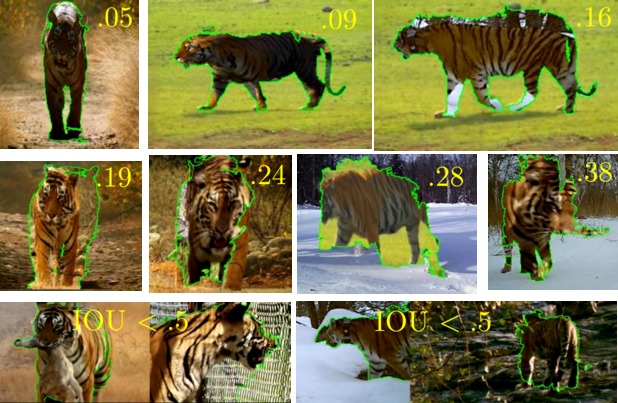}
\end{center}
 \caption{\small{
   Alignment error. We use the ground-truth landmarks
   to measure the alignment error of the mappings estimated by our method (sec.~\ref{sec:alignmenterror}).
   As the error increases, the quality of the alignment clearly degrades.  Around $0.18$ the alignments contain
   some slight mistakes (\eg, the slightly misaligned legs in the top right image),
   but are typically acceptable. We consider an alignment incorrect when the error is above
   $0.18$, and also when the IOU of the 
   visible landmarks in the aligned pair is below $0.5$ (bottom row).
   }}
\label{fig:reprojectionerror}
\end{figure}

After visual inspection of many sampled alignments \linebreak (fig.~\ref{fig:reprojectionerror}), we found that $0.18$ was a 
reasonable threshold for separating acceptable alignments from those with noticeable errors.
We count an alignment as correct if the error is below this threshold and if the Intersection-over-Union (IOU) of the two sets of visible landmarks in the sequences is 
above $0.5$~\footnote{If $\mathcal{L}_{1}$ is the set of landmarks visible in the first sequence in a CMP,
and $\mathcal{L}_{2}$ those in the second, IoU$(\mathcal{L}_{1},\mathcal{L}_{2})=|\mathcal{L}_{1} \cap \mathcal{L}_{2}| / |\mathcal{L}_{1} \cup \mathcal{L}_{2}|$. For example, if $\mathcal{L}_{1}$=\{left\textunderscore eye,right\textunderscore eye,neck\}
and $\mathcal{L}_{2}$=\{front\textunderscore right\textunderscore knee, right\textunderscore shoulder, neck\}, IoU=$1/5$ }. 
This prevents rewarding accidental alignments of a few landmarks (bottom row of fig.~\ref{fig:reprojectionerror}).

\subsubsection{Results on CMP extraction}
\label{sec:cmpevaluation}
First, we evaluate our method for CMP extraction 
in isolation (sec.~\ref{sec:cmpcandidates}). Given a CMP, 
we fit a homography to correspondences between the ground-truth
landmarks, and check if it is correct based on the alignment error above.
This indicates that it is possible to align the CMP (we call it \emph{alignable}).
Computing (\ref{eq:cmpcandidates}) using both PoTs and MBH returns roughly $3,000$ CMP on tigers, 
of which $51\%$ are alignable ($43\%$ if we use only PoTs).
As a baseline, we extract CMPs directly from the input shots:
we select the starting frames of the two sequences in a CMP by sampling 
from a uniform distribution over all input frames (\ie without
step 2 and step 3 in fig.~\ref{fig:architecture}). 
The percentage of alignable CMPs produced by this baseline is only $19\%$.
Results are similar on horses: our method delivers $49\%$ alignable CMPs
($47\%$ using only PoTs), vs.\ $26\%$ by the baseline.

\subsubsection{Results on spatial alignment}
\label{sec:resultsalignment}
We now evaluate our methods for sequence alignment (sec.~\ref{sec:homography} and~\ref{sec:tps}).
For each, we generate a precision-recall curve as follows.
Let $n$ be the total number of CMPs returned by the method; $c$ the number of correctly aligned CMPs; and $a$ the total number of alignable CMPs (sec.~\ref{sec:cmpevaluation}).
Recall is $c/a$, and precision is $c/n$.
Different operating points on the precision-recall curve are obtained by varying the 
maximum percentage of outliers allowed when fitting a homography.

\paragraph{Baselines.}
We compare our method against SIFT Flow~\cite{liu08eccv}.
We use SIFT Flow to align each pair of frames from the
two CMP sequences independently.
We help the SIFT Flow algorithm by matching only the bounding boxes of the foreground masks,
after rescaling them to be the same size. Without these two stjpg, the algorithm fails on most CMPs.

We also compare to fitting a homography to SIFT matches between the two sequences.
We use only keypoints on the foreground mask, and preserve
temporal order by matching only keypoints in corresponding frames.
We tested this method alone (SIFT), and by adding spatial regularization  
with the foreground masks (SIFT + FG, as in sec.~\ref{sec:homographyregularized}).
Finally, we consider a simple baseline that fits a homography to the 
bounding boxes of the foreground masks alone (FG).

\begin{figure}
\begin{center}
\includegraphics[scale = 0.23]{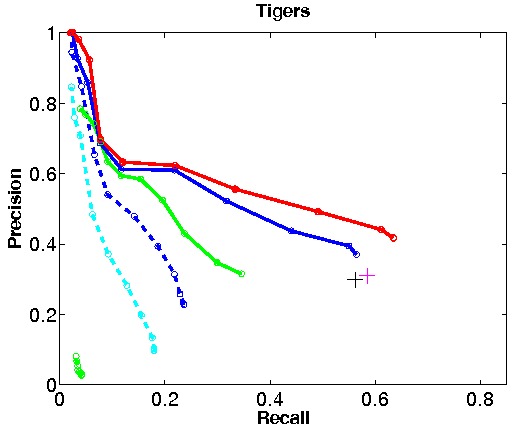}
\includegraphics[scale = 0.23]{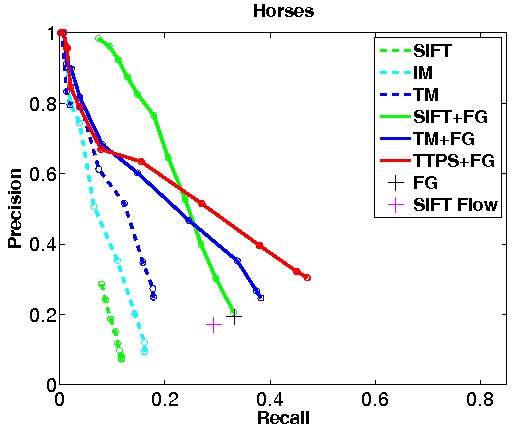}
\end{center}
 \caption{\small{
Evaluation of sequence alignment. We separately evaluate our method
on two classes, horses and tigers (sec.~\ref{sec:resultsalignment}).
With no regularization, 
trajectory methods are superior to SIFT on both classes,
with TM performing better than IM. 
Adding regularization using the foreground masks (+FG)
improves the performance of both TM and SIFT (compare
the dashed to the solid curves). 
TTPS clearly outperform all trajectory methods,
as well as SIFT Flow and the FG baseline (sec.~\ref{sec:resultsalignment}). }}
\label{fig:alignmentresults}
\end{figure}

\begin{figure}
\begin{center}
\includegraphics[scale = 0.47]{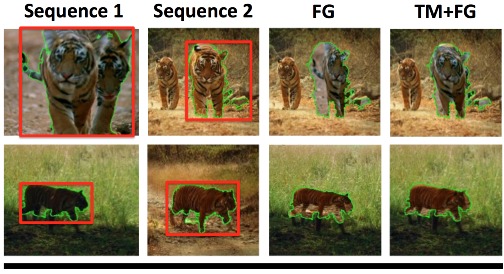}
\includegraphics[scale = 0.47]{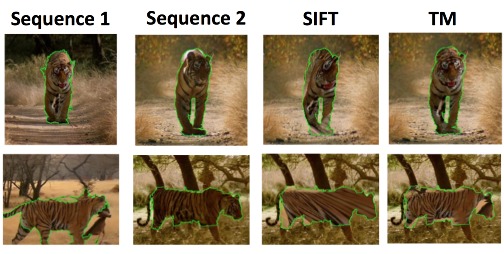}
\end{center}
 \caption{\small{
 Top two rows: Estimating the homography from the foreground masks alone (FG) fails
 when the bounding boxes are not tight around the objects (first-second columns). 
 Adding trajectories (TM+FG) is more accurate (sec.~\ref{sec:homographyregularized}).
 Bottom two rows: the striped texture of tigers often confuses estimating
 the homography from SIFT keypoint matches (third column). 
 On this class, using trajectories (TM) often performs better (sec.~\ref{sec:resultsalignment}).
 }}
\label{fig:alignmentcomparison}
\end{figure}

\begin{figure}
\begin{center}
\setlength{\tabcolsep}{0.8pt}
\begin{tabular}{c c c c}
\footnotesize{\textbf{Sequence 1}} & \footnotesize{\textbf{Sequence 2}} & \footnotesize{\textbf{Homography}} & \footnotesize{\textbf{TTPS}} \\
\includegraphics[scale =0.15]{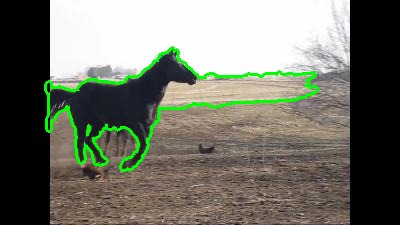}  &
\includegraphics[scale =0.15]{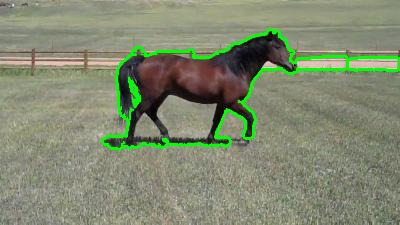}  &
\includegraphics[scale =0.15]{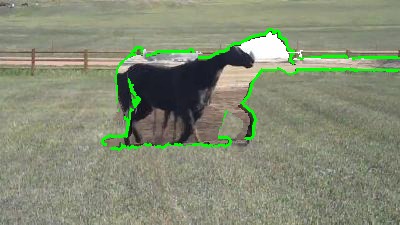}  &
\includegraphics[scale =0.15]{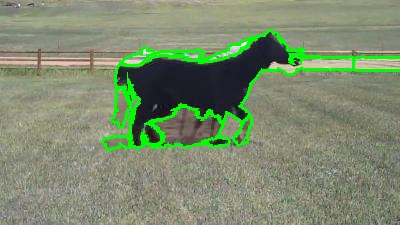} \\
\includegraphics[scale =0.15]{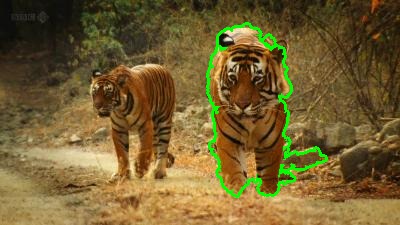}  &
\includegraphics[scale =0.15]{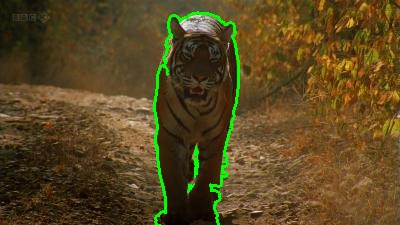}  &
\includegraphics[scale =0.15]{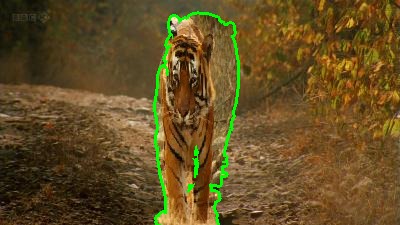}  &
\includegraphics[scale =0.15]{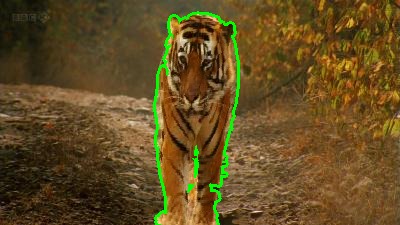} \\
\small{(a)} & \small{(b)} & \small{(c)} & \small{(d)} \\
\end{tabular}
\end{center}
 \caption{\small{
TTPS (d) provide a more accurate alignment for complex articulated objects than
homographies (c, sec~\ref{sec:resultsalignment}). 
%A video showing many sequences aligned by our method
%is available at~\cite{delpero15cvpr-potswebpage}
}}
\label{fig:alignmentexamples} 
\end{figure}

\paragraph{}We report results in fig.~\ref{fig:alignmentresults}. Among the homography-based methods (sec.~\ref{sec:homography}),
those using trajectory correspondences (TM, IM, sec.~\ref{sec:homographyvideos}) are superior to using SIFT on both classes, with TM outperforming IM.
Adding spatial regularization with the foreground masks (+FG) improves the performance of both TM and SIFT.
SIFT performs poorly on tigers, since the striped texture confuses matching SIFT
keypoints (fig.~\ref{fig:alignmentcomparison}, bottom).
Methods using trajectories work somewhat better on tigers than horses due to the poorer quality of YouTube video (\eg low resolution, shaky camera, abrupt pans).
As a consequence, TM+FG clearly outperforms SIFT+FG on tigers, but it is somewhat worse
on horses.

The time-varying TPS model (TTPS+FG, sec.~\ref{sec:tps}) significantly improves upon its initialization (TM+FG) on both classes. % Science Bit
% now the fray, kill the competitors
On tigers, it is the best method overall, as its performance curve is above all others for the entire range.
On horses, the SIFT+FG and TTPS+FG curves intersect. However, TTPS+FG achieves a higher Average Precision (\ie the area under the curve): 0.265 vs.\ 0.235.

The SIFT Flow software~\cite{liu08eccv} does not produce scores comparable across CMPs, so we cannot produce a full \linebreak precision-recall curve. At the level of recall of SIFT Flow, TTPS+FG achieves +0.2 higher precision on tigers, and +0.3 on horses. We also note that TM and TM+FG are closely 
related to the method for fitting homographies
to trajectories in~\cite{caspi06ijcv}.
Although TM+FG extends~\cite{caspi06ijcv} in several ways (automatic CMP extraction, modified TS descriptor, regularization with foreground masks), it is still inferior to TTPS+FG.
Last, TTPS+FG also achieves a significantly higher precision than the FG baseline. 
This shows that our method is robust to errors in the foreground masks (fig.~\ref{fig:alignmentcomparison}, top).
Head-to-head qualitative results show that TTPS+FG alignments typically look more accurate 
than the other methods (fig.~\ref{fig:alignmentexamples}). A video with many examples
is available on our website~\cite{delpero15cvpr-potswebpage}.

For the tiger class, out of all CPMs returned by TTPS+FG (rightmost point on the curve),
$1,000$ of them are correctly aligned (\ie $10,000$ frames). 
The precision at this point is $0.5$, \ie half of the returned CMPs are correctly aligned.
For the horse class, TTPS+FG returns 800 correctly aligned CMPs, with precision $0.35$.

\begin{table}
\begin{center}
\small
\setlength{\tabcolsep}{20pt}
\begin{tabular}{lc}
\toprule
\textbf{step} & \textbf{run-time} \tabularnewline
\midrule
optical flow~\cite{brox11pami} (per frame) & 1.5s \tabularnewline
\hline
foreground mask (per frame) & 0.5s \tabularnewline
\hline
dense trajectory extraction (per frame) & 0.4s \tabularnewline
\hline
PoT extraction (per frame) & 0.1s \tabularnewline
\hline
homography alignment (per CMP) & 5s \tabularnewline
\hline
TTPS alignment (per CMP) & 44s \tabularnewline
\hline
\end{tabular}
\end{center}
 \caption{\small
Run-time of the main stjpg of our method (sec.~\ref{sec:runtime}).
}
 \label{table:runtime}
\end{table}

\subsection{Runtime}
\label{sec:runtime}
We report the run-time of the main stjpg of our method in Table~\ref{table:runtime}, including pre-processing. 
We measured run-time on a Dell server with a 1.6 GHz CPU
and 16GB RAM. The PoT extraction time is negligible compared to the
pre-processing stjpg (optical flow, foreground mask and dense trajectory
extraction). We note that large video collections can be processed efficiently
on a computer cluster,
since each input shot (or CMP for the alignment) can be processed independently.

\subsection{Analysis of failures and limitations}
\label{sec:limitations}
\begin{figure}
\begin{center}
\includegraphics[scale = 0.3]{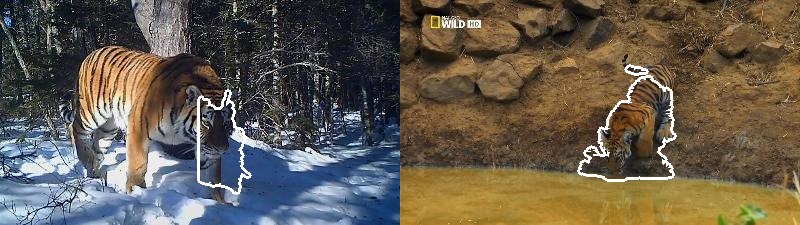}
\end{center}
 \caption{\small{
Failures due to inaccurate foreground mask. Our system is robust to inaccuracies in the foreground 
masks (sec.~\ref{sec:timeclustering} and fig.~\ref{fig:tpsedges}), but cannot recover
when the the object is almost completely missed (left).
Here the walking tiger (left) was clustered with the tiger sitting 
down (right) during behavior discovery (sec.~\ref{sec:timeclustering}). This in turn breaks the alignment stage,
as these two tigers cannot be aligned via homography or TTPS (sec.~\ref{sec:alignment}).
We estimated by visual inspection that complete failures in the masks happen
in roughly $15\%$ of the input shots (sec.~\ref{sec:limitations}).
   }}
\label{fig:failure}
\end{figure}

\paragraph{Inaccurate foreground masks.}
Our system is robust to small to medium inaccuracies in the foreground 
masks, such as missing part of the object or including some of the background
(see sec.~\ref{sec:timeclustering} and fig.~\ref{fig:tpsedges}).
However, we cannot cope with catastrophic
failures, for example when the object is completely missed.
In these cases the PoT extraction is not reliable, which results in assigning such shots
to the wrong behavior cluster (fig.~\ref{fig:failure}), which in turn 
produces wrong alignments in the following step of our system.
However, these problematic cases are not frequent (about
$15\%$ of the input shots).
%and can be assigned to a cluster corresponding to a completely
%different behavior (which in turn results in incorrect alignment in the following step).
Moreover, we noticed that hierarchical clustering often puts such an item in a singleton cluster,
which mitigates the problem. 
Inaccuracies in the masks can potentially be detected and fixed by co-segmenting all the intervals 
in a behavior cluster, while
enforcing consistent appearance and shape across all their foreground masks. 
%This in turn might allow to refine the alignment stage, which is also sensitive 
%to significant inaccuracies in the segmentation masks.
%
\paragraph{Scale and viewpoint invariance.} The PoT descriptor is invariant to scale
(sec.~\ref{sec:potdefinition}).
In general, smaller objects will generate fewer trajectories (hence fewer PoTs), but this is not a
problem since we aggregate the PoTs into a normalized BOW histogram (sec.~\ref{sec:timeclustering}).
Our results show that our method clusters together objects at a very different scale (\eg
fig.~\ref{fig:qualitative}b).
Only cases where the object
is very small are problematic ($<50 \times 50$ pixels).
PoTs are also robust
to moderate viewpoint and pose variations. However,
they cannot cope with drastic viewpoint difference, \eg a video of a tiger walking frontally and one walking to the right. Establishing correspondences between clusters showing the same
behavior under widely
different viewpoints is an interesting research direction. 
%For example, one could enforce a consistent mapping between each video in the first cluster to each video in the second.
%
\paragraph{Camera motion.} The PoT descriptor can cope with camera panning, 
and other moderate camera motions (sec.~\ref{sec:potdefinition}).
The foreground masks also help in the presence of panning, since the motion of
rigid regions of the object and the background would be indistinguishable in this case.
However, fast zooming can be problematic.
\paragraph{Extensions to multiple classes.} The main goal of our system is to organize a collection
of videos of the same class. However, extensions to multiple classes are possible.
In the case of related classes (\eg quadrupeds),
similar behaviors of different classes might be grouped together, and 
additional cues might be needed to separate them.

\section{Discussion}
\label{sec:discussion}
We introduced a weakly supervised system that discovers the behaviors
of an articulated object class from unconstrained video, while also spatially aligning several instances of each behavior.
We emphasize that the only supervision needed is a single label per video, indicating which class it contains.

The entire system is bottom-up and needs not relate to the kinematic structure of
an object class.
We showed that the behavior discovery and the alignment process apply to different
classes, by leveraging the recurring motion patterns of a particular class, rather than being limited to 
pre-defined relationships.

This was enabled by our PoT descriptor, which proves very effective
for modeling the motion of articulated objects. 
Thanks to the use of pairs of trajectories, PoTs outperform
alternative motion descriptors (\eg TS) on behavior discovery.
While being appearance-free, on horses and tigers PoTs also outperform all tested alternatives that included 
appearance information (\eg IDTFs).
When augmented with appearance descriptors, PoTs also outperforms competitors on the dog class.
In terms of spatial alignment, we have shown that our technique produces more accurate alignments than relevant alternatives such as SIFT Flow and SIFT matching.

Thanks to the principled use of motion, we discovered behaviors and recovered alignments
across instances exhibiting significant appearance variations (orange and white
tigers, cubs and adults, etc.). Establishing such correspondences across different object instances
can be very useful to learn class-level models of behavior and/or appearance.
Our method recovers them automatically from unconstrained Internet video,
and can be a platform for replacing the tedious and expensive manual
annotations normally needed when learning from video.

%An essential feature of our method is that a collection of PoTs 
%can encode detailed information about the 
%relative motion between many different parts of an object.
%PoT anchors are scattered across the
%object; each may move with its own unique trajectory.
%Simplifying PoTs to a star-like model 
%where all anchors coincide with the center of mass of the
%object (\ie, normalizing by the dominant object motion) would
%result in a loss of expressive power and would be less robust for highly
%deformable objects.

%We have shown that clustering built on top of PoTs
%finds motion patterns that are consistent across many
%shots. While many common behaviors (\eg, walking) are cyclic, our method
%focuses instead on consistency across occurrences 
%rather than periodicity within an occurrence.
%Periodic motion is exploited during partitioning, but the clustering procedure
%itself makes no such assumption, enabling us to discover behaviors such as a tiger turning its head.

\begin{acknowledgements}
We are very grateful to Anestis Papazoglou
for helping with the data collection, and to Shumeet
Baluja for his helpful comments. This work was partly
funded by a Google Faculty Research Award, and by ERC
Starting Grant ``Visual Culture for Image Understanding".
We also thank the reviewers for their helpful comments.
\end{acknowledgements}

\bibliographystyle{spmpsci}      % mathematics and physical sciences
\bibliography{./longstrings,./calvin,./vggroup}   % name your BibTeX data base

%\begin{thebibliography}{}
% etc
%\end{thebibliography}

\end{document}